\documentclass[conference]{IEEEtran}
\IEEEoverridecommandlockouts

\PassOptionsToPackage{table,xcdraw,dvipsnames}{xcolor}

\usepackage{algorithmic}
\usepackage{amsmath,amssymb,amsfonts}
\usepackage{booktabs}
\usepackage{cite}
\usepackage{colortbl}
\usepackage{graphicx}
\usepackage{multirow}
\usepackage{pgfplots}
\pgfplotsset{compat=1.18}
\usepackage{subcaption}
\usepackage{tabularx}
\usepackage{textcomp}
\usepackage{xcolor}

\def\BibTeX{{\rm B\kern-.05em{\sc i\kern-.025em b}\kern-.08em
    T\kern-.1667em\lower.7ex\hbox{E}\kern-.125emX}}
\begin{document}

\title{Clinically-Informed Modeling for Pediatric Brain Tumor Classification from Whole-Slide Histopathology Images}

\author{
\IEEEauthorblockN{Joakim Nguyen\textsuperscript{†}}
\IEEEauthorblockA{\textit{Dept. of Computer Science} \\
\textit{University of Texas at Austin}\\
Austin, TX, USA \\
jhn001@utexas.edu}
\and
\IEEEauthorblockN{Jian Yu\textsuperscript{†}}
\IEEEauthorblockA{\textit{Dept. of Computer Science} \\
\textit{University of Texas at Austin}\\
Austin, TX, USA \\
jian.yu@utexas.edu}
\and
\IEEEauthorblockN{Jinrui Fang}
\IEEEauthorblockA{\textit{School of Information} \\
\textit{University of Texas at Austin}\\
Austin, TX, USA \\
jinrui@utexas.edu}
\and
\IEEEauthorblockN{Nicholas Konz}
\IEEEauthorblockA{\textit{Dept. of Computer Science} \\
\textit{UNC at Chapel Hill}\\
Chapel Hill, NC, USA \\
nickk124@cs.unc.edu}
\and
\IEEEauthorblockN{Tianlong Chen}
\IEEEauthorblockA{\textit{Dept. of Computer Science} \\
\textit{UNC at Chapel Hill}\\
Chapel Hill, NC, USA \\
tianlong@cs.unc.edu}
\and
\IEEEauthorblockN{Sanjay Krishnan}
\IEEEauthorblockA{\textit{Dept. of Pathology} \\
\textit{Dell Children's Medical Center}\\
Austin, TX, USA \\
snjkris2009@gmail.com}
\and
\IEEEauthorblockN{Chandra Krishnan}
\IEEEauthorblockA{\textit{Dept. of Pathology} \\
\textit{Dell Children's Medical Center}\\
Austin, TX, USA \\
c.krishnan@thinkoculus.com}
\and
\IEEEauthorblockN{Ying Ding}
\IEEEauthorblockA{\textit{School of Information} \\
\textit{University of Texas at Austin}\\
Austin, TX, USA \\
ying.ding@ischool.utexas.edu}
\and
\IEEEauthorblockN{\phantom{Name}}
\IEEEauthorblockA{\phantom{Affiliation}}
\and
\IEEEauthorblockN{Hairong Wang\textsuperscript{‡}}
\IEEEauthorblockA{\textit{Dept. of OREI} \\
\textit{University of Texas at Austin}\\
Austin, TX, USA \\
hairong@utexas.edu}
\and
\IEEEauthorblockN{Ankita Shukla\textsuperscript{‡}}
\IEEEauthorblockA{\textit{Dept. of Comptuer Science and Engineering} \\
\textit{University of Nevada, Reno}\\
Reno, NV, USA \\
ankitas@unr.edu}
\and
\IEEEauthorblockN{\phantom{Name}}
\IEEEauthorblockA{\phantom{Affiliation}}
\thanks{\textsuperscript{†}Equal contribution. \textsuperscript{‡}Co-corresponding authors.}
}

\maketitle

\begin{abstract}
Accurate diagnosis of pediatric brain tumors, starting with histopathology, presents unique challenges for deep learning, including severe data scarcity, class imbalance, and fine-grained morphologic overlap  across diagnostically distinct subtypes. While pathology foundation models have advanced patch-level representation learning, their effective adaptation to weakly supervised pediatric brain tumor classification under limited data remains underexplored. In this work, we introduce an expert-guided contrastive fine-tuning framework for pediatric brain tumor diagnosis from whole-slide images (WSI). Our approach integrates contrastive learning into slide-level multiple instance learning (MIL) to explicitly regularize the geometry of slide-level representations during downstream fine-tuning. We propose both a general supervised contrastive setting and an expert-guided variant that incorporates clinically informed hard negatives targeting diagnostically confusable subtypes. Through comprehensive experiments on pediatric brain tumor WSI classification under realistic low-sample and class-imbalanced conditions, we demonstrate that contrastive fine-tuning yields measurable improvements in fine-grained diagnostic distinctions. Our experimental analyses reveal complementary strengths across different contrastive strategies, with expert-guided hard negatives promoting more compact intra-class representations and improved inter-class separation. This work highlights the importance of explicitly shaping slide-level representations for robust fine-grained classification in data-scarce pediatric pathology settings.

\end{abstract}


\section{Introduction}




Applying deep learning to pediatric brain tumor diagnosis poses challenges that are fundamentally different from those encountered in adult oncology \cite{price2025cbtrus}. 
Pediatric brain tumors are rare by nature, resulting in limited cohort sizes even at large pediatric centers. 
In contrast to common adult malignancies such as breast or lung cancer, where multi-institutional datasets with tens of thousands of cases are routinely available, pediatric brain tumor studies are often constrained to a few hundred patients at best. 
This limitation is further compounded by the expansive biological and histopathological heterogeneity of pediatric central nervous system tumors defined under the current WHO classification \cite{louis20212021}. The advent of molecular genetics and methylation based tumor profiling continues to refine existing diagnostic categories as well as creating new diagnostic entities. 
Considering the rarity of pediatric brain tumors and the expanding diagnostic classifications that exist,  many clinically relevant subtypes are represented by only a small number of cases within institutional datasets, creating substantial barriers to training deep learning models that generalize reliably in fine-grained diagnostic settings.

Accurate and timely diagnosis of pediatric brain tumors from histopathology is critical for guiding treatment decisions and improving clinical outcomes, as integrated histologic and molecular classification directly informs risk stratification and therapeutic planning  \cite{tampu2026pediatric, viaene2023pediatric}. However, compared to adult neuro-oncology, pediatric brain tumor pathology presents unique challenges for data-driven modeling: clinically meaningful subtypes are often rare, the tumor types encountered in children are different and the frequency of these classes vary greatly by age, and many distinct pediatric brain tumor classes show significant histopathologic overlap on H\&E slides  \cite{tampu2025pediatric,dorfner2025review,louis20212021}. These factors substantially hinder generalization, especially in multi-class, fine-grained diagnostic settings.

While deep learning has achieved substantial success in medical image analysis, many pediatric brain tumor studies have predominantly focused on radiological modalities such as MRI \cite{tampu2025pediatric}. 
In contrast, computational pathology approaches based on whole-slide images (WSIs) remain relatively underexplored, largely due to the scarcity of pediatric pathology datasets and the high cost of expert-driven data curation and annotation \cite{tampuPediatricBrainTumor2024a,tampuPediatricBrainTumor2025}. Due to the gigapixel resolution, WSIs are typically partitioned into a large number of image patches for computational analysis. However, detailed annotations at the patch level are rarely available in practice. To address this limitation, weakly supervised learning has become a natural paradigm for WSI analysis, where slide-level labels are used to train models over sets of image patches through multiple instance learning (MIL) \cite{zhangPatchesWSIsSystematic2024,ilse2018attention}. 
Recent work has begun to investigate weakly supervised pediatric brain tumor classification using digital histopathology, often leveraging MIL combined with strong patch-level representations \cite{tampuPediatricBrainTumor2024a,tampuPediatricBrainTumor2025}. 
Nevertheless, fine-grained subtype recognition under data scarcity and class imbalance remains an open challenge, and prior studies in adult glioma pathology suggest that achieving robust model adaptation remains challenging, even in more data-rich settings \cite{saueressigHistologyDiagnosisLeveraging2025}.

In parallel, the emergence of pathology foundation models has substantially advanced representation learning in computational pathology. 
Large-scale self-supervised pretraining has produced transferable patch-level features that improve performance across a wide range of downstream pathology tasks \cite{chenGeneralpurposeFoundationModel2023,karasikovTrainingStateoftheartPathology2025}. 
One representative pathology foundation model is UNI (Universal Pathology Image Foundation Model), which exemplifies the recent trend toward large-scale self-supervised pretraining on diverse histopathology data. Recent surveys and benchmarking studies further highlight both the promise and current limitations of pathology foundation models, emphasizing the importance of data diversity, model design, and effective adaptation under weak supervision \cite{xiongSurveyPathologyFoundation2025,neidlingerBenchmarkingFoundationModels2024,vadoriMindGapEvaluating2025}. Despite these advances, most progress has been driven by improving patch-level representations or MIL aggregation architectures, while comparatively less attention has been devoted to explicitly regularizing the geometry of slide-level representations during downstream MIL fine-tuning, particularly in low-sample, fine-grained diagnostic regimes.

In this work, we address this gap by introducing \textbf{expert-guided contrastive learning (EGCL)} framework for weakly supervised pediatric brain tumor diagnosis from WSIs. 
Our approach builds on a strong pathology foundation model and a slide-level MIL pipeline, and focuses on shaping slide-level representations during downstream fine-tuning under data scarcity. Specifically, we systematically study contrastive learning (CL) in the MIL fine-tuning stage, comparing a generic supervised contrastive learning setting to an expert-guided design that integrates clinically informed hard negatives. 
By explicitly targeting diagnostically confusable subtypes, the proposed framework promotes more compact intra-class representations and improved inter-class separation at the slide level. We evaluate our method on pediatric brain tumor WSI classification under realistic low-sample and class-imbalanced conditions, and perform comprehensive analyses to assess improvements in fine-grained recognition and representation structure across diagnostic categories.

Our contributions are as follows:
\begin{itemize}
  \item We propose a contrastive fine-tuning framework for weakly supervised pediatric brain tumor diagnosis from WSI that integrates contrastive learning into slide-level MIL, and explicitly studies how contrastive supervision at the slide level shapes representation geometry under data scarcity. The framework includes both a general supervised contrastive setting and an expert-guided variant that incorporates clinical knowledge via diagnostically confusable hard negatives, facilitating fine-grained multi-class separation.

  \item We conduct comprehensive experiments and analyses in pediatric brain tumor pathology to systematically compare foundation-model-based MIL baselines, general supervised contrastive fine-tuning, and expert-guided hard-negative designs. Our results reveal complementary strengths across settings, with different approaches excelling in different diagnostic scenarios. We further provide detailed error analyses that explain these behaviors from both a representation learning perspective and a clinical interpretation standpoint.

\end{itemize}

\begin{figure*}
    \centering
    \includegraphics[width=\linewidth]{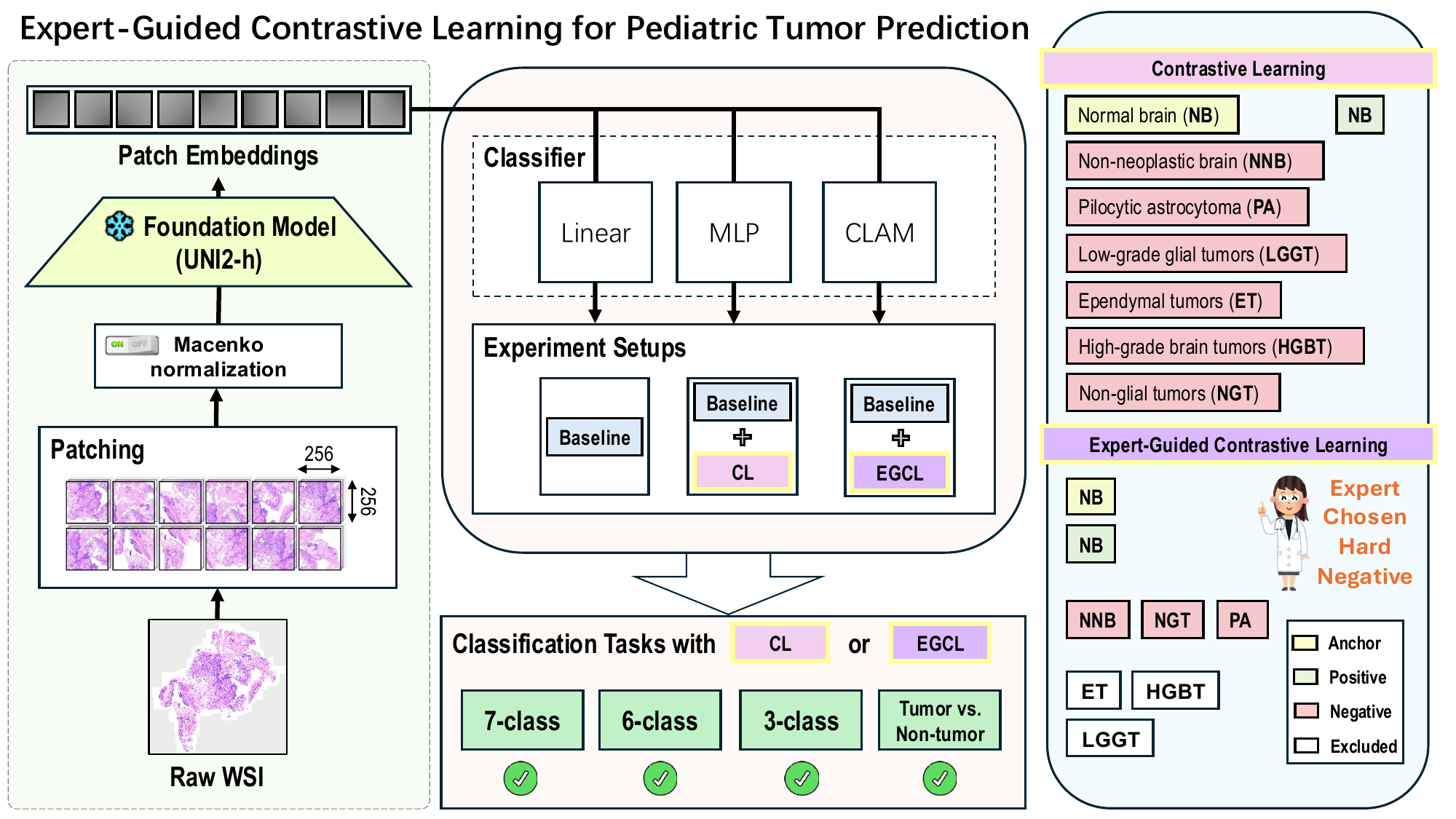}
    \caption{Overview of our slide-level modeling pipeline and experimental configurations for pediatric brain tumor prediction. (\emph{Left}) A whole-slide image (WSI) is partitioned into patches, each patch is encoded using a frozen pathology foundation model, and tile embeddings are aggregated by a multiple instance learning (MIL) slide encoder to produce a single slide representation used for diagnosis. (\emph{Middle}) We train and compare a grid of settings that vary stain normalization (Macenko on/off) and supervision objective: cross-entropy only (Baseline) versus adding a contrastive loss at the slide-representation level (CL) to shape the embedding geometry during downstream fine-tuning. (\emph{Right}) In expert-guided contrastive learning (EGCL), the contrastive objective is further specialized by restricting/weighting negative pairs to expert-defined hard-negative categories corresponding to clinically confusable tumor subtypes, encouraging separation of diagnostically similar classes. Results are reported under multiple label granularities (2-, 3-, 6-, and 7-class tasks).}
    \label{fig:Figure 1}
\end{figure*}

\section{Related Work}

Computational pathology for pediatric brain tumor diagnosis remains relatively underdeveloped compared to adult neuro-oncology, largely due to limited data availability, pronounced histopathological heterogeneity, and clinically meaningful fine-grained subtypes. 
Recent progress in this area has been driven by advances along three complementary directions: deep learning methods tailored to pediatric brain tumor diagnosis, weakly supervised learning frameworks for WSI analysis, and self-supervised representation learning in computational pathology.
In this section, we review related work across these three dimensions to contextualize recent developments and to highlight the remaining challenges at the intersection of data scarcity, weak supervision, and fine-grained diagnostic tasks.

\subsection{Deep Learning for Pediatric Brain Tumor Diagnosis} 

UNI is a representative example, trained via large-scale self-supervised learning on histopathology patches spanning diverse organs, stains, and species, and shown to improve weakly supervised WSI analysis when combined with downstream MIL pipelines \cite{chenGeneralpurposeFoundationModel2023,karasikovTrainingStateoftheartPathology2025}.

Most existing deep learning studies on pediatric brain tumors have predominantly focused on radiological modalities such as MRI \cite{tampu2025pediatric}. For example, recent work has explored MRI-based tumor classification by incorporating auxiliary clinical information, such as patient age, to improve diagnostic performance in pediatric cohorts \cite{tampu2025pediatric}. While these approaches demonstrate the value of multimodal cues in pediatric neuro-oncology, they remain limited to specific imaging pipelines and do not address pathology-based diagnosis. In contrast, pathology-based approaches using WSIs remain relatively underexplored, largely due to the scarcity of pediatric pathology datasets and the high cost of expert-driven data curation and annotation \cite{tampuPediatricBrainTumor2024a,tampuPediatricBrainTumor2025}. 

Recent work has begun to investigate weakly supervised learning for pediatric brain tumor classification using digital histopathology, leveraging multiple instance learning (MIL) and pathology foundation models to address annotation constraints. These studies demonstrate the promise of computational pathology in pediatric neuro-oncology, but primarily emphasize hierarchical or coarse-grained diagnostic tasks. 

Despite these advances, fine-grained subtype classification under severe data scarcity and class imbalance remains an open challenge. Moreover, evidence from adult glioma studies suggests that while pathology foundation models can yield strong downstream performance, robustness and effective adaptation across datasets remain non-trivial \cite{saueressigHistologyDiagnosisLeveraging2025}.

\subsection{Multiple Instance Learning and Weak Supervision for WSIs}
Multiple instance learning (MIL) has emerged as the dominant paradigm for WSI analysis in computational pathology, enabling slide-level diagnosis without requiring patch-level annotations. Comprehensive surveys have documented a wide range of MIL formulations, including pooling-based, attention-based, and clustering-constrained approaches, highlighting their effectiveness across diverse pathology tasks \cite{zhangPatchesWSIsSystematic2024,ilse2018attention}.

Beyond standard attention-based MIL, numerous extensions have been proposed to improve aggregation and contextual modeling. These include transformer-based MIL architectures that incorporate spatial or positional encoding to capture inter-patch relationships \cite{shiPositionalEncodingguidedTransformerbased2024}, as well as sequential attention-based sampling strategies designed to improve computational efficiency by selectively analyzing informative regions \cite{gSequentialAttentionbasedSampling2025}.

Recent studies have also questioned the necessity of complex MIL architectures for slide-level fine-tuning, suggesting that simpler pooling-based strategies can be competitive when combined with strong patch-level representations \cite{liCanWeSimplify2025}. Together, these works underscore that representation quality and aggregation strategy are both critical components of weakly supervised WSI analysis.

However, despite architectural diversity, MIL-based approaches are typically optimized using bag-level classification objectives and often provide limited explicit control over the geometry of slide-level representations \cite{zhangPatchesWSIsSystematic2024}. 
This limitation becomes particularly pronounced in fine-grained multi-class settings with severe class imbalance, where consistent separation of diagnostically similar subtypes remains challenging.

\subsection{Contrastive Learning in Computational Pathology}



Contrastive learning has become a cornerstone of self-supervised representation learning, with frameworks such as MoCo and SimCLR enabling scalable learning of discriminative features from unlabeled data \cite{he2020momentumcontrastunsupervisedvisual, chen2020simple}.
In medical imaging, including computational pathology, contrastive learning is widely adopted as a pretraining strategy to learn transferable representations from large-scale datasets such as WSIs, giving rise to pathology foundation models with strong downstream generalization \cite{chenGeneralpurposeFoundationModel2023,xiongSurveyPathologyFoundation2025}.
UNI is a representative example, trained via large-scale self-supervised learning on histopathology patches spanning diverse organs, stains, and species, and shown to improve weakly supervised WSI analysis when combined with downstream MIL pipelines \cite{chenGeneralpurposeFoundationModel2023,karasikovTrainingStateoftheartPathology2025}.

Recent surveys and benchmarking studies show that strong pretraining alone is insufficient to address downstream challenges, highlighting the need for effective adaptation under weak supervision \cite{xiongSurveyPathologyFoundation2025,neidlingerBenchmarkingFoundationModels2024,vadoriMindGapEvaluating2025}.
Vision–language foundation models extend this paradigm by integrating histopathology with clinical or textual information and demonstrate promise in large-scale oncology applications \cite{dingMultimodalWholeslideFoundation2025,xiangVisionLanguageFoundation2025}.
However, these approaches primarily focus on large-scale representation learning and pretraining, and leave open problem of how to effectively adapt pretrained representations at the slide level under weak supervision in low-sample, fine-grained diagnostic settings.

While contrastive learning is widely used during self-supervised pretraining, its role in downstream fine-tuning has received limited attention.
In data-scarce regimes, contrastive objectives can regularize representation geometry by encouraging intra-class compactness and inter-class separability, yet their use in MIL-based slide-level fine-tuning remains underexplored \cite{xiongSurveyPathologyFoundation2025}.
This gap is particularly critical for pediatric brain tumor pathology, where subtle morphological differences and limited data exacerbate the difficulty of multi-class discrimination.

\section{Dataset}
\begin{figure}
    \centering
    \includegraphics[width=\linewidth]{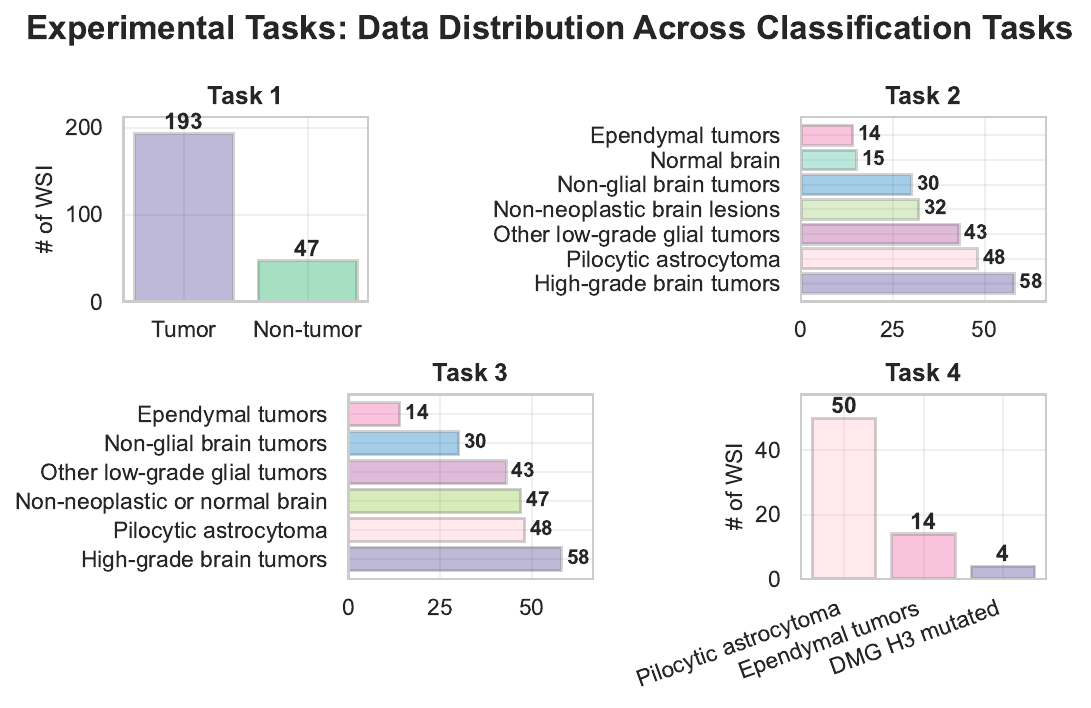}
    \caption{Overview of the four slide-level classification tasks derived from our pediatric brain WSI cohort, illustrating how diagnostic categories are grouped across the 2-, 3-, 6-, and 7-class settings and the corresponding per-class slide counts.}
    \label{fig:task_datasets}
\end{figure}

This study utilized a de-identified institutional dataset collected under IRB approval \textbf{STUDY00007710} at Dell Children’s Medical Center. The dataset comprises 237 pediatric brain WSIs (Hematoxylin and Eosin stained), spanning seven categories (five tumor classes and two control/non-tumor classes). These WSIs were drawn from an institutional database containing records for 292 patients. Each WSI is associated with structured metadata including patient age, tumor location, histopathologic diagnosis,  molecular and immunohistochemical findings. Genetic annotations capture the presence or absence of key gene alterations, while immunostaining results inform the presence or absence of expression of certain key proteins found in brain tumors (  \textbf{GFAP}, \textbf{Synaptophysin}, \textbf{INI-1}, \textbf{H3K27M}, and \textbf{ALK-1)}. Each record also includes a categorical \textbf{methylation classification label}, serving as a proxy for molecular subtype, where applicable.

WSIs were processed with a standard digital pathology pipeline to produce patch-level inputs for feature extraction and downstream slide-level modeling; full segmentation, tiling, and stain-normalization details are provided in Figure~\ref{fig:Figure 1} and the Methods section. All analyses were conducted on de-identified data in compliance with institutional ethical standards.

We defined four slide-level classification tasks from the diagnostic categories in our cohort:

\textbf{Task 1: Binary Classification (2 classes)} - Tumor vs. non-tumor discrimination, serving as a baseline diagnostic task distinguishing neoplastic from non-neoplastic tissue.

\textbf{Task 2: 7-Class Classification} - Comprehensive classification across the full diagnostic spectrum: Ependymal tumors, High-grade brain tumors, Non-glial brain tumors, Non-neoplastic brain lesions, Normal brain, Other low-grade glial tumors, and Pilocytic astrocytoma.

\textbf{Task 3: 6-Class Classification} - Classification that merges Non-neoplastic brain lesions and Normal brain into a single control class while keeping tumor categories distinct. We include this setting so the model does not over-focus on separating Normal vs. Non-neoplastic controls; instead, it prioritizes learning tumor-class boundaries while still leveraging both control tissue types.

\textbf{Task 4: 3-Class Classification} - A fine-grained differentiation among three clinically important pediatric brain tumor subtypes: DMG H3 mutated, Pilocytic astrocytoma, and Ependymal tumors.

Validation is currently limited to a single institution, as available public datasets (e.g., TCGA) lack slide-level pediatric labels aligned with our 3-, 6-, and 7-class definitions, preventing direct cross-cohort benchmarking. Multi-institutional pediatric cohorts with harmonized label schemas are a key future priority.

\section{Methods}
\noindent An overview of proposed framework is shown in Figure \ref{fig:Figure 1}. 
\subsection{Segmentation and Tiling Processing setup}

As a first step to process the WSIs, we perform tissue segmentation and patching.  The segmentation was performed using the TRIDENT library with the \texttt{hest} segmentation method and a segmentation confidence threshold of 0.5~\cite{zhang2025standardizing,vaidya2025molecular,jaume2024hest}. Following segmentation, WSIs were tiled at 20$\times$ magnification into 256$\times$256 pixel patches with 0 pixel overlap. In addition to the native (non-normalized) patches, we generated a parallel dataset in which Macenko stain normalization was applied patch-wise to each extracted RGB patch prior to feature extraction. Macenko normalization reduces stain-related color variability by converting RGB to optical density, estimating per-slide stain vectors, and linearly mapping each patch to a fixed reference stain basis~\cite{Macenko2009AMF}.
We use these Macenko-normalized patches only as an ablation to evaluate the sensitivity of downstream models to stain variation

\subsection{Pathology Foundation Models}
For patch-level feature extraction, we leverage the UNI2-h model (ViT-H/14) as a frozen backbone. UNI2-h is a pathology foundation model trained on more than 200 million histological image patches using DINOv2 self-supervised learning, making it one of the largest ViT-based encoders for H\&E images. We input 256×256 pixel tissue patches (at 20× magnification) into UNI2-h and use the resulting [CLS] token embedding (1536 dimensions) as the patch feature vector. We also evaluated whether applying Macenko stain normalization to patches would impact the foundation model features. Consistent with recent studies~\cite{saueressigHistologyDiagnosisLeveraging2025}, we found that UNI’s patch representations are robust to color variations and the model’s performance remained essentially unchanged with or without Macenko normalization, indicating strong invariance to stain differences.


\subsection{Classification Architecture}
For slide-level classification, we adopt the \textbf{multi-branch CLAM} (Clustering-constrained Attention MIL) architecture. Multi-branch CLAM uses class-specific attention-based pooling to identify the most informative instances for each class and imposes a clustering constraint to improve the discriminability of high-attention patches~\cite{lu2021clam}. As illustrated in \textbf{Fig.~\ref{fig:Figure 1}}, the UNI2-h backbone acts as a frozen feature extractor, and all downstream classification and contrastive objectives operate on \emph{processed} UNI features.

Given a patch $\mathbf{x}_{i,p}$ from slide $i$, UNI produces a patch embedding (the [CLS] token):
\begin{equation}
    \mathbf{u}_{i,p} = f_{\text{UNI}}(\mathbf{x}_{i,p}) \in \mathbb{R}^{1536}.
\end{equation}

These UNI embeddings are then mapped to the CLAM feature space via a learnable linear projection:
\begin{equation}
    \mathbf{h}_{i,p} = g_{\phi}(\mathbf{u}_{i,p}) \in \mathbb{R}^{d},
\end{equation}
which serves as input to attention pooling and the slide-level classifier. Let $a_{i,p,c}$ denote the multi-branch CLAM attention weight assigned to patch $p$ for class $c$ in slide $i$, computed by an attention network and normalized with a softmax over patches (for each class). Multi-branch CLAM forms a class-specific slide (bag) representation by attention-weighted aggregation:
\begin{equation}
    \mathbf{r}_{i,c} = \sum_{p=1}^{P_i} a_{i,p,c}\,\mathbf{h}_{i,p}, \quad \sum_{p=1}^{P_i} a_{i,p,c}=1.
\end{equation}
For contrastive learning, we use the embedding corresponding to the ground-truth label, $\mathbf{r}_{i,y_i}$, as the bag feature. The slide-level classifier then outputs class probabilities by applying a per-class linear head to $\mathbf{r}_{i,c}$ and normalizing across classes:
\begin{equation}
    \hat{\mathbf{y}}_i = \mathrm{softmax}\big([c_{\psi}(\mathbf{r}_{i,1}),\dots,c_{\psi}(\mathbf{r}_{i,C})]\big).
\end{equation}

For the contrastive objective, we apply $\mathcal{L}_{\text{CL}}$ directly to the CLAM bag feature after $\ell_2$ normalization:
\begin{equation}
    \mathbf{z}_i = \mathrm{norm}(\mathbf{r}_{i,y_i}).
\end{equation}
Importantly, $\mathcal{L}_{\text{CL}}$ is computed on $\mathbf{z}_i$ (i.e., features \emph{after} extraction from UNI and \emph{after} CLAM aggregation), not directly on the raw UNI patch outputs $\mathbf{u}_{i,p}$.

CLAM produces interpretable attention scores while jointly learning slide-level classification with instance-level feature regularization. The total loss is defined as:

\begin{equation}
    \mathcal{L}_{\text{total}} = \mathcal{L}_{\text{bag}} + \mathcal{L}_{\text{inst}} + \lambda\mathcal{L}_{\text{CL}} + \alpha \|\theta\|_2^2
\end{equation}

\noindent where $\mathcal{L}_{\text{bag}}$ is the cross-entropy loss for slide-level classification, $\mathcal{L}_{\text{inst}}$ is the instance-level SVM loss for attention supervision, $\lambda$ is the contrastive loss weight, $\alpha = 10^{-5}$ is the L2 regularization weight, and $\theta$ represents the model parameters.

We incorporate a queue-based supervised contrastive loss during training to encourage separation of slide (bag) embeddings in feature space. For each anchor slide $i$, we contrast its normalized embedding $\mathbf{z}_i$ against a memory queue $\mathcal{Q}_i$ of slide embeddings stored from recent batches. The loss follows an InfoNCE-style formulation~\cite{oord2018representation}, adapted for supervised CL with a memory queue mechanism~\cite{khosla2020supervised}:

\begin{equation}
    \mathcal{L}_{\text{CL}} = -\frac{1}{N}\sum_{i=1}^{N} \frac{1}{|P(i)|} \sum_{j \in P(i)} \log \frac{\exp(s_{ij}/\tau)}{\sum_{k \in \mathcal{Q}_i} \exp(s_{ik}/\tau)}
\end{equation}

\noindent where $s_{ij} = \text{sim}(\mathbf{z}_i, \mathbf{z}_j) = \mathbf{z}_i \cdot \mathbf{z}_j / (\|\mathbf{z}_i\| \|\mathbf{z}_j\|)$ is the cosine similarity between normalized embeddings, $P(i) = \{j \in \mathcal{Q}_i : y_j = y_i\}$ denotes positive samples in the queue (same class), and $\tau$ is a temperature parameter ($\tau=0.1$ in our experiments). In practice, we compute $\mathcal{L}_{\text{CL}}$ only when at least one positive exists in the queue, and we maintain a fixed-size memory queue of 256 embeddings to provide a diverse set of negatives across batches.

\subsection{Expert-Guided Negative Sampling}

Furthermore, we experiment with our CL variant, \textbf{EGCL}, where expert-defined hard negatives are specified as clinically confusable class pairs (Normal$\leftrightarrow$Non-neoplastic, Non-neoplastic$\leftrightarrow\{$Other LGG, Non-glial$\}$, Other LGG$\leftrightarrow$Pilocytic astrocytoma, Pilocytic astrocytoma$\leftrightarrow$Ependymal, and Ependymal$\leftrightarrow$High-grade), determined a priori by neuropathology review, and incorporated during training by up-weighting these near-miss negatives in the queue-based contrastive denominator (all other negatives retained with unit weight).

To isolate the effect of contrastive learning, we additionally evaluate two simpler MIL baselines: (1)~\textbf{MeanMIL~+~Linear}, which averages patch-level embeddings per slide and applies a single linear classifier, and (2)~\textbf{MeanMIL~+~MLP}, which uses a 2-layer multilayer perceptron on mean-pooled features. These baselines lack attention mechanisms, providing a controlled setting to assess the impact of contrastive learning without instance-level selection. We emphasize that the primary goal of this work is not exhaustive benchmarking of all WSI architectures, but evaluation of whether incorporating expert-guided class-similarity priors improves representation geometry in a pediatric brain tumor setting where publicly benchmarked alternatives remain limited. All models are trained on frozen UNI2-h patch features.

\section{Experimental Setup and Results}
\newcommand{\BlockTitle}[1]{%
  \par\noindent\rule{\textwidth}{0.45pt}\par
  \vspace{0.8pt}
  \noindent\makebox[\textwidth][c]{\scriptsize #1}\par
  \vspace{0.8pt}
  \vspace{3pt}
}

\begin{table*}[h]
\centering

\caption{Patient-stratified 10-fold cross-validation results for pediatric brain tumor WSI classification across 7-/6-/3-/2-class tasks. We report per-class precision (P), recall (R), and F1-score (F1) and macro averages for Baseline, supervised contrastive learning (CL), and expert-guided contrastive learning (EGCL) under different MIL heads (Linear/MLP/CLAM) and with/without Macenko stain normalization. \textbf{Bold} indicates the best performance, and \underline{underlined} indicates the second best.}
\label{tab:crossval_results}

\BlockTitle{\textit{7-Class Classification}}

\resizebox{\textwidth}{!}{
\begin{tabular}{@{} l l ccc ccc ccc ccc ccc ccc ccc | ccc @{}}
\toprule
\multirow{2}{*}{\textbf{Method}} &
\multirow{2}{*}{\textbf{Model}} &
\multicolumn{3}{c}{Ependymal tumors} &
\multicolumn{3}{c}{High-grade brain tumors} &
\multicolumn{3}{c}{Non-glial brain tumors} &
\multicolumn{3}{c}{Non-neoplastic brain lesions} &
\multicolumn{3}{c}{Normal brain} &
\multicolumn{3}{c}{Other low-grade glial tumors} &
\multicolumn{3}{c|}{Pilocytic astrocytoma} &
\multicolumn{3}{c}{Macro} \\
\cmidrule(lr){3-5}\cmidrule(lr){6-8}\cmidrule(lr){9-11}\cmidrule(lr){12-14}\cmidrule(lr){15-17}\cmidrule(lr){18-20}\cmidrule(lr){21-23}\cmidrule(lr){24-26}
 &  & P & R & F1 & P & R & F1 & P & R & F1 & P & R & F1 & P & R & F1 & P & R & F1 & P & R & F1 & P & R & F1 \\
\midrule
\multirow{3}{*}{Baseline} &
Linear & \underline{0.857} & \underline{0.600} & 0.706 & 0.794 & 0.833 & 0.813 & 0.667 & 0.667 & 0.667 & 0.667 & \textbf{0.600} & \underline{0.632} & \textbf{1.000} & 0.765 & 0.867 & 0.578 & 0.650 & 0.612 & 0.814 & 0.854 & 0.833 & \underline{0.768} & 0.710 & 0.733 \\

& MLP & 0.700 & \textbf{0.700} & 0.700 & 0.791 & \textbf{0.883} & 0.835 & 0.643 & 0.600 & 0.621 & 0.654 & \underline{0.567} & 0.607 & \textbf{1.000} & 0.706 & 0.828 & 0.590 & 0.575 & 0.582 & 0.761 & 0.854 & 0.805 & 0.734 & 0.698 & 0.711 \\

& CLAM & \textbf{0.875} & \textbf{0.700} & \textbf{0.778} & 0.810 & 0.850 & 0.829 & 0.704 & 0.633 & 0.667 & 0.667 & 0.533 & 0.593 & 0.813 & 0.765 & 0.788 & 0.578 & 0.650 & 0.612 & 0.733 & 0.805 & 0.767 & 0.740 & 0.705 & 0.719 \\
\midrule

\multirow{3}{*}{CL} &
Linear & 0.750 & \underline{0.600} & 0.667 & 0.785 & 0.850 & 0.816 & \textbf{0.818} & 0.600 & 0.692 & 0.654 & \underline{0.567} & 0.607 & \textbf{1.000} & 0.706 & 0.828 & 0.549 & \underline{0.700} & 0.615 & \underline{0.818} & 0.878 & 0.847 & \underline{0.768} & 0.700 & 0.725 \\
& MLP & 0.700 & \textbf{0.700} & 0.700 & 0.788 & \underline{0.867} & 0.825 & 0.690 & 0.667 & 0.678 & 0.667 & 0.533 & 0.593 & 0.929 & 0.765 & 0.839 & 0.595 & 0.625 & 0.610 & 0.814 & 0.854 & 0.833 & 0.740 & 0.716 & 0.725 \\
& CLAM & \textbf{0.875} & \textbf{0.700} & \textbf{0.778} & \textbf{0.850} & 0.850 & \underline{0.850} & 0.733 & \textbf{0.733} & \textbf{0.733} & \textbf{0.773} & \underline{0.567} & \textbf{0.654} & \underline{0.944} & \textbf{1.000} & \textbf{0.971} & \textbf{0.683} & \underline{0.700} & \textbf{0.691} & 0.694 & 0.829 & 0.756 & \textbf{0.793} & \textbf{0.768} & \textbf{0.776} \\
\midrule

\multirow{3}{*}{EGCL} &
Linear & 0.750 & \underline{0.600} & 0.667 & 0.781 & 0.833 & 0.806 & \underline{0.769} & 0.667 & \underline{0.714} & 0.654 & \underline{0.567} & 0.607 & \textbf{1.000} & 0.647 & 0.786 & 0.571 & \underline{0.700} & 0.629 & \underline{0.818} & 0.878 & 0.847 & 0.763 & 0.699 & 0.722 \\
& MLP & 0.700 & \textbf{0.700} & 0.700 & 0.794 & 0.833 & 0.813 & 0.714 & 0.667 & 0.690 & \underline{0.680} & \underline{0.567} & 0.618 & \textbf{1.000} & 0.765 & 0.867 & 0.543 & 0.625 & 0.581 & 0.814 & 0.854 & 0.833 & 0.749 & 0.716 & 0.729 \\
& CLAM & \textbf{0.875} & \textbf{0.700} & \textbf{0.778} & 0.823 & 0.850 & 0.836 & 0.724 & \underline{0.700} & 0.712 & 0.667 & \textbf{0.600} & \underline{0.632} & 0.929 & 0.765 & 0.839 & 0.628 & 0.675 & 0.651 & 0.733 & 0.805 & 0.767 & \underline{0.768} & 0.728 & \underline{0.745} \\
\midrule

\multirow{3}{*}{Macenko} &
Linear & 0.714 & 0.500 & 0.588 & 0.823 & 0.850 & 0.836 & 0.633 & 0.633 & 0.633 & 0.577 & 0.500 & 0.536 & 0.917 & 0.647 & 0.759 & 0.595 & 0.625 & 0.610 & 0.776 & \textbf{0.927} & 0.844 & 0.719 & 0.669 & 0.687 \\
& MLP & 0.571 & 0.400 & 0.471 & 0.810 & 0.850 & 0.829 & 0.655 & 0.633 & 0.644 & 0.583 & 0.467 & 0.519 & 0.923 & 0.706 & 0.800 & 0.575 & 0.575 & 0.575 & 0.731 & \textbf{0.927} & 0.817 & 0.693 & 0.651 & 0.665 \\
& CLAM & 0.778 & \textbf{0.700} & \underline{0.737} & 0.825 & \underline{0.867} & 0.846 & 0.667 & 0.600 & 0.632 & 0.600 & 0.500 & 0.545 & 0.846 & 0.647 & 0.733 & 0.636 & \underline{0.700} & 0.667 & 0.787 & \underline{0.902} & 0.841 & 0.734 & 0.702 & 0.714 \\
\midrule

\multirow{3}{*}{Macenko + CL} &
Linear & 0.750 & \underline{0.600} & 0.667 & 0.839 & \underline{0.867} & \textbf{0.852} & 0.692 & 0.600 & 0.643 & 0.625 & 0.500 & 0.556 & 0.857 & 0.706 & 0.774 & 0.596 & \underline{0.700} & 0.644 & 0.809 & \textbf{0.927} & \underline{0.864} & 0.738 & 0.700 & 0.714 \\
& MLP & 0.600 & \underline{0.600} & 0.600 & 0.817 & 0.817 & 0.817 & 0.667 & 0.667 & 0.667 & 0.583 & 0.467 & 0.519 & 0.923 & 0.706 & 0.800 & 0.575 & 0.575 & 0.575 & 0.725 & \underline{0.902} & 0.804 & 0.699 & 0.676 & 0.683 \\
& CLAM & 0.700 & \textbf{0.700} & 0.700 & \underline{0.842} & 0.800 & 0.821 & 0.667 & 0.667 & 0.667 & 0.625 & 0.500 & 0.556 & 0.875 & 0.824 & 0.848 & 0.659 & \textbf{0.725} & \underline{0.690} & 0.766 & 0.878 & 0.818 & 0.733 & 0.728 & 0.729 \\
\midrule

\multirow{3}{*}{Macenko + EGCL} &
Linear & 0.750 & \underline{0.600} & 0.667 & 0.823 & 0.850 & 0.836 & 0.643 & 0.600 & 0.621 & 0.600 & 0.500 & 0.545 & 0.857 & 0.706 & 0.774 & 0.600 & 0.675 & 0.635 & \textbf{0.826} & \textbf{0.927} & \textbf{0.874} & 0.728 & 0.694 & 0.707 \\
& MLP & 0.500 & 0.500 & 0.500 & 0.817 & 0.817 & 0.817 & 0.655 & 0.633 & 0.644 & 0.600 & 0.500 & 0.545 & \textbf{1.000} & 0.706 & 0.828 & 0.571 & 0.600 & 0.585 & 0.740 & \underline{0.902} & 0.813 & 0.698 & 0.665 & 0.676 \\
& CLAM & 0.700 & \textbf{0.700} & 0.700 & \underline{0.842} & 0.800 & 0.821 & 0.625 & 0.667 & 0.645 & 0.652 & 0.500 & 0.566 & 0.882 & \underline{0.882} & \underline{0.882} & \underline{0.667} & \underline{0.700} & 0.683 & 0.787 & \underline{0.902} & 0.841 & 0.737 & \underline{0.736} & 0.734 \\
\bottomrule
\end{tabular}

}

\vspace{0.9em}

\BlockTitle{\textit{6-Class Classification}}

\resizebox{\textwidth}{!}{

\begin{tabular}{@{} l l ccc ccc ccc ccc ccc ccc |ccc @{}}
\toprule
\multirow{2}{*}{\textbf{Method}} &
\multirow{2}{*}{\textbf{Model}} &
\multicolumn{3}{c}{Ependymal tumors} &
\multicolumn{3}{c}{High-grade brain tumors} &
\multicolumn{3}{c}{Non-glial brain tumors} &
\multicolumn{3}{c}{Non-neoplastic / normal brain} &
\multicolumn{3}{c}{Other low-grade glial tumors} &
\multicolumn{3}{c|}{Pilocytic astrocytoma} &
\multicolumn{3}{c}{Macro} \\
\cmidrule(lr){3-5}\cmidrule(lr){6-8}\cmidrule(lr){9-11}\cmidrule(lr){12-14}\cmidrule(lr){15-17}\cmidrule(lr){18-20}\cmidrule(lr){21-23}
 &  & P & R & F1 & P & R & F1 & P & R & F1 & P & R & F1 & P & R & F1 & P & R & F1 & P & R & F1 \\
\midrule

\multirow{3}{*}{Baseline} &
Linear & 0.818 & \textbf{0.900} & \underline{0.857} & \textbf{0.864} & \textbf{0.850} & \textbf{0.857} & 0.750 & \textbf{0.700} & \textbf{0.724} & \underline{0.818} & \underline{0.750} & \textbf{0.783} & 0.553 & 0.500 & 0.525 & 0.686 & 0.854 & 0.761 & 0.748 & 0.759 & 0.751 \\
& MLP   & 0.875 & 0.700 & 0.778 & 0.797 & \textbf{0.850} & 0.823 & 0.750 & 0.600 & 0.667 & 0.787 & \textbf{0.771} & \underline{0.779} & 0.514 & 0.452 & 0.481 & 0.667 & 0.829 & 0.739 & 0.732 & 0.700 & 0.711 \\
& CLAM  & \textbf{0.900} & \textbf{0.900} & \textbf{0.900} & 0.806 & \underline{0.833} & 0.820 & 0.720 & 0.600 & 0.655 & \textbf{0.833} & 0.729 & 0.778 & 0.463 & 0.452 & 0.458 & 0.667 & 0.829 & 0.739 & 0.732 & 0.724 & 0.725 \\
\midrule

\multirow{3}{*}{CL} &
Linear & \textbf{0.900} & \textbf{0.900} & \textbf{0.900} & 0.823 & \textbf{0.850} & 0.836 & \underline{0.760} & 0.633 & 0.691 & 0.778 & 0.729 & 0.753 & 0.571 & \underline{0.571} & \underline{0.571} & \underline{0.745} & 0.854 & \underline{0.795} & \underline{0.763} & 0.756 & \underline{0.758} \\
& MLP   & 0.800 & \underline{0.800} & 0.800 & 0.823 & \textbf{0.850} & 0.836 & 0.731 & 0.633 & 0.679 & 0.783 & \underline{0.750} & 0.766 & 0.537 & 0.524 & 0.530 & 0.717 & 0.805 & 0.759 & 0.732 & 0.727 & 0.728 \\
& CLAM  & 0.818 & \textbf{0.900} & \underline{0.857} & \underline{0.833} & \underline{0.833} & 0.833 & 0.696 & 0.533 & 0.604 & 0.778 & 0.729 & 0.753 & 0.556 & 0.476 & 0.513 & 0.625 & 0.854 & 0.722 & 0.718 & 0.721 & 0.714 \\
\midrule

\multirow{3}{*}{EGCL} &
Linear & \textbf{0.900} & \textbf{0.900} & \textbf{0.900} & 0.823 & \textbf{0.850} & 0.836 & 0.714 & \underline{0.667} & 0.690 & \underline{0.800} & \underline{0.750} & 0.774 & \textbf{0.615} & \underline{0.571} & \textbf{0.593} & \underline{0.745} & 0.854 & \underline{0.795} & \textbf{0.766} & \textbf{0.765} & \textbf{0.765} \\
& MLP   & 0.800 & \underline{0.800} & 0.800 & 0.810 & \textbf{0.850} & \underline{0.829} & 0.720 & 0.600 & 0.655 & 0.783 & \underline{0.750} & 0.766 & 0.513 & 0.476 & 0.494 & 0.688 & 0.805 & 0.742 & 0.719 & 0.714 & 0.714 \\
& CLAM  & \textbf{0.900} & \textbf{0.900} & \textbf{0.900} & 0.820 & \underline{0.833} & 0.826 & 0.720 & 0.600 & 0.655 & 0.814 & 0.729 & 0.769 & 0.529 & 0.429 & 0.474 & 0.621 & \textbf{0.878} & 0.727 & 0.734 & 0.728 & 0.725 \\
\midrule

\multirow{3}{*}{Macenko} &
Linear & 0.818 & \textbf{0.900} & \underline{0.857} & \underline{0.833} & \underline{0.833} & 0.833 & 0.690 & \underline{0.667} & 0.678 & 0.778 & 0.729 & 0.753 & 0.571 & 0.476 & 0.519 & 0.686 & 0.854 & 0.761 & 0.729 & 0.743 & 0.734 \\
& MLP   & \textbf{0.900} & \textbf{0.900} & \textbf{0.900} & \underline{0.862} & \underline{0.833} & \underline{0.847} & 0.630 & 0.567 & 0.596 & 0.745 & 0.729 & 0.737 & 0.524 & 0.524 & 0.524 & 0.702 & 0.805 & 0.750 & 0.727 & 0.726 & 0.726 \\
& CLAM  & 0.818 & \textbf{0.900} & \underline{0.857} & 0.817 & 0.817 & 0.817 & 0.667 & 0.533 & 0.593 & \underline{0.800} & 0.667 & 0.727 & 0.532 & \textbf{0.595} & \underline{0.562} & 0.714 & 0.854 & 0.778 & 0.725 & 0.728 & 0.722 \\
\midrule

\multirow{3}{*}{Macenko + CL} &
Linear & 0.818 & \textbf{0.900} & \underline{0.857} & 0.850 & \textbf{0.850} & \textbf{0.850} & \textbf{0.792} & 0.633 & \underline{0.704} & 0.783 & \underline{0.750} & 0.766 & 0.590 & 0.548 & 0.568 & 0.725 & \underline{0.902} & \textbf{0.804} & \underline{0.760} & \underline{0.764} & \underline{0.758} \\
& MLP   & \underline{0.889} & \underline{0.800} & 0.842 & 0.810 & \textbf{0.850} & \underline{0.829} & 0.654 & 0.567 & 0.607 & 0.739 & 0.708 & 0.723 & 0.558 & \underline{0.571} & 0.565 & \textbf{0.750} & 0.805 & 0.776 & 0.733 & 0.717 & 0.724 \\
& CLAM  & \textbf{0.900} & \textbf{0.900} & \textbf{0.900} & 0.754 & 0.817 & 0.784 & 0.667 & 0.467 & 0.549 & 0.767 & 0.688 & 0.725 & 0.561 & 0.548 & 0.554 & 0.686 & 0.854 & 0.761 & 0.723 & 0.712 & 0.712 \\
\midrule

\multirow{3}{*}{Macenko + EGCL} &
Linear & 0.750 & \textbf{0.900} & 0.818 & 0.836 & \textbf{0.850} & 0.843 & 0.741 & \underline{0.667} & 0.702 & \underline{0.800} & \underline{0.750} & 0.774 & \underline{0.600} & 0.500 & 0.545 & 0.725 & \underline{0.902} & \textbf{0.804} & 0.742 & 0.762 & 0.748 \\
& MLP   & \textbf{0.900} & \textbf{0.900} & \textbf{0.900} & 0.820 & \underline{0.833} & 0.826 & 0.640 & 0.533 & 0.582 & 0.733 & 0.688 & 0.710 & 0.561 & 0.548 & 0.554 & 0.714 & 0.854 & 0.778 & 0.728 & 0.726 & 0.725 \\
& CLAM  & \textbf{0.900} & \textbf{0.900} & \textbf{0.900} & 0.817 & 0.817 & 0.817 & 0.680 & 0.567 & 0.618 & 0.762 & 0.667 & 0.711 & 0.533 & \underline{0.571} & 0.552 & 0.694 & 0.829 & 0.756 & 0.731 & 0.725 & 0.726 \\
\bottomrule
\end{tabular}

}

\vspace{0.9em}

\BlockTitle{\textit{3-Class Classification \& Binary Classification}}
\resizebox{\textwidth}{!}{
\begin{tabular}{@{} l l ccc ccc ccc | ccc || ccc ccc | ccc @{}}
\toprule
\multirow{2}{*}{\textbf{Method}} &
\multirow{2}{*}{\textbf{Model}} &
\multicolumn{3}{c}{DMG H3 mutated} &
\multicolumn{3}{c}{Pilocytic astrocytoma} &
\multicolumn{3}{c|}{Ependymal tumors} &
\multicolumn{3}{c||}{Macro (3-class)} &
\multicolumn{3}{c}{Non-tumor} &
\multicolumn{3}{c|}{Tumor} &
\multicolumn{3}{c}{Macro (2-class)} \\
\cmidrule(lr){3-5}\cmidrule(lr){6-8}\cmidrule(lr){9-11}\cmidrule(lr){12-14}\cmidrule(lr){15-17}\cmidrule(lr){18-20}\cmidrule(lr){21-23}
 & & P & R & F1 & P & R & F1 & P & R & F1 & P & R & F1 & P & R & F1 & P & R & F1 & P & R & F1 \\
\midrule

\multirow{3}{*}{Baseline} &
Linear & \textbf{1.000} & \underline{0.200} & \underline{0.333} & 0.870 & \textbf{1.000} & 0.930 & 0.583 & \textbf{0.700} & 0.636 & 0.818 & \underline{0.633} & 0.633 & 0.900 & 0.804 & 0.849 & 0.945 & 0.974 & 0.959 & 0.922 & 0.889 & 0.904 \\
& MLP   & \textbf{1.000} & \underline{0.200} & \underline{0.333} & \textbf{0.909} & \textbf{1.000} & \textbf{0.952} & 0.500 & \textbf{0.700} & 0.583 & 0.803 & \underline{0.633} & 0.623 & 0.900 & 0.804 & 0.849 & 0.945 & 0.974 & 0.959 & 0.922 & 0.889 & 0.904 \\
& CLAM  & \underline{0.667} & \underline{0.200} & 0.308 & 0.867 & \underline{0.975} & 0.918 & 0.583 & \textbf{0.700} & 0.636 & 0.706 & 0.625 & 0.621 & 0.833 & 0.804 & 0.818 & 0.944 & 0.953 & 0.948 & 0.888 & 0.878 & 0.883 \\
\midrule

\multirow{3}{*}{CL} &
Linear & \textbf{1.000} & \underline{0.200} & \underline{0.333} & 0.870 & \textbf{1.000} & 0.930 & 0.583 & \textbf{0.700} & 0.636 & 0.818 & \underline{0.633} & 0.633 & 0.692 & 0.804 & 0.744 & 0.940 & 0.896 & 0.918 & 0.816 & 0.850 & 0.831 \\
& MLP   & 0.500 & \underline{0.200} & 0.286 & \underline{0.902} & 0.925 & 0.914 & 0.467 & \textbf{0.700} & 0.560 & 0.623 & 0.608 & 0.586 & 0.902 & \underline{0.821} & 0.860 & 0.949 & 0.974 & 0.962 & 0.926 & 0.898 & 0.911 \\
& CLAM  & \textbf{1.000} & \textbf{0.300} & \textbf{0.462} & 0.848 & \underline{0.975} & 0.907 & 0.636 & \textbf{0.700} & 0.667 & 0.828 & \textbf{0.658} & \textbf{0.678} & 0.807 & \underline{0.821} & 0.814 & 0.948 & 0.943 & 0.945 & 0.877 & 0.882 & 0.880 \\
\midrule

\multirow{3}{*}{Macenko} &
Linear & \textbf{1.000} & \underline{0.200} & \underline{0.333} & 0.816 & \textbf{1.000} & 0.899 & \underline{0.778} & \textbf{0.700} & \underline{0.737} & \underline{0.865} & \underline{0.633} & 0.656 & 0.849 & 0.804 & 0.826 & 0.944 & 0.959 & 0.951 & 0.896 & 0.881 & 0.888 \\
& MLP   & \textbf{1.000} & \underline{0.200} & \underline{0.333} & 0.870 & \textbf{1.000} & 0.930 & 0.583 & \textbf{0.700} & 0.636 & 0.818 & \underline{0.633} & 0.633 & 0.849 & 0.804 & 0.826 & 0.944 & 0.959 & 0.951 & 0.896 & 0.881 & 0.888 \\
& CLAM  & \textbf{1.000} & \underline{0.200} & \underline{0.333} & 0.889 & \textbf{1.000} & \underline{0.941} & 0.538 & \textbf{0.700} & 0.609 & 0.809 & \underline{0.633} & 0.628 & \textbf{0.959} & \textbf{0.839} & \textbf{0.895} & \textbf{0.955} & \textbf{0.990} & \textbf{0.972} & \textbf{0.957} & \textbf{0.914} & \textbf{0.934} \\
\midrule

\multirow{3}{*}{Macenko + CL} &
Linear & \textbf{1.000} & \underline{0.200} & \underline{0.333} & 0.800 & \textbf{1.000} & 0.889 & \textbf{0.875} & \textbf{0.700} & \textbf{0.778} & \textbf{0.892} & \underline{0.633} & \underline{0.667} & 0.839 & \textbf{0.839} & 0.839 & \underline{0.953} & 0.953 & 0.953 & 0.896 & 0.896 & 0.896 \\
& MLP   & \textbf{1.000} & \underline{0.200} & \underline{0.333} & 0.870 & \textbf{1.000} & 0.930 & 0.583 & \textbf{0.700} & 0.636 & 0.818 & \underline{0.633} & 0.633 & 0.852 & \underline{0.821} & 0.836 & 0.949 & 0.959 & 0.954 & 0.900 & 0.890 & 0.895 \\
& CLAM  & \underline{0.667} & \underline{0.200} & 0.308 & 0.851 & \textbf{1.000} & 0.920 & 0.700 & \textbf{0.700} & 0.700 & 0.739 & \underline{0.633} & 0.642 & \underline{0.922} & \textbf{0.839} & \underline{0.879} & \textbf{0.955} & \underline{0.979} & \underline{0.967} & \underline{0.938} & \underline{0.909} & \underline{0.923} \\
\bottomrule
\end{tabular}

}

\end{table*}

\subsection{Training Protocol}

To ensure robust evaluation and prevent data leakage, we used patient-stratified 10-fold cross-validation with data split at the patient level rather than the slide level, so all slides from a given patient appeared exclusively in one split. Each fold used 80\%/10\%/10\% train/validation/test partitions, with at least one available sample per class in validation and test when feasible for small class counts. When a patient had multiple slides, patient-level labels were determined by majority vote over slide predictions. The random seed was set to 42 for splitting and training to ensure reproducibility.


\subsection{Evaluation Metrics}

Model performance was evaluated using comprehensive metrics appropriate for multi-class medical image classification:

\begin{itemize}
    \item \textbf{Per-Class Metrics}: For each diagnostic class, we computed precision, recall, and F1-score to assess class-specific discrimination ability.
    
    \item \textbf{Aggregate Metrics}: Overall classification accuracy, macro-averaged precision, recall, and F1-score were calculated to summarize global performance across all classes. Additionally, weighted F1-score was computed to account for class imbalance.
    
    \item \textbf{Cross-Validation Analysis}: Metrics were computed independently for each fold and then aggregated across all folds to provide both fold-specific and overall performance estimates with associated variability measures.
\end{itemize}

Results were logged systematically, with confusion matrices generated to visualize classification patterns and identify systematic misclassifications between specific diagnostic categories.

\begin{figure*}[h!]
    \centering
    \includegraphics[width=\linewidth]{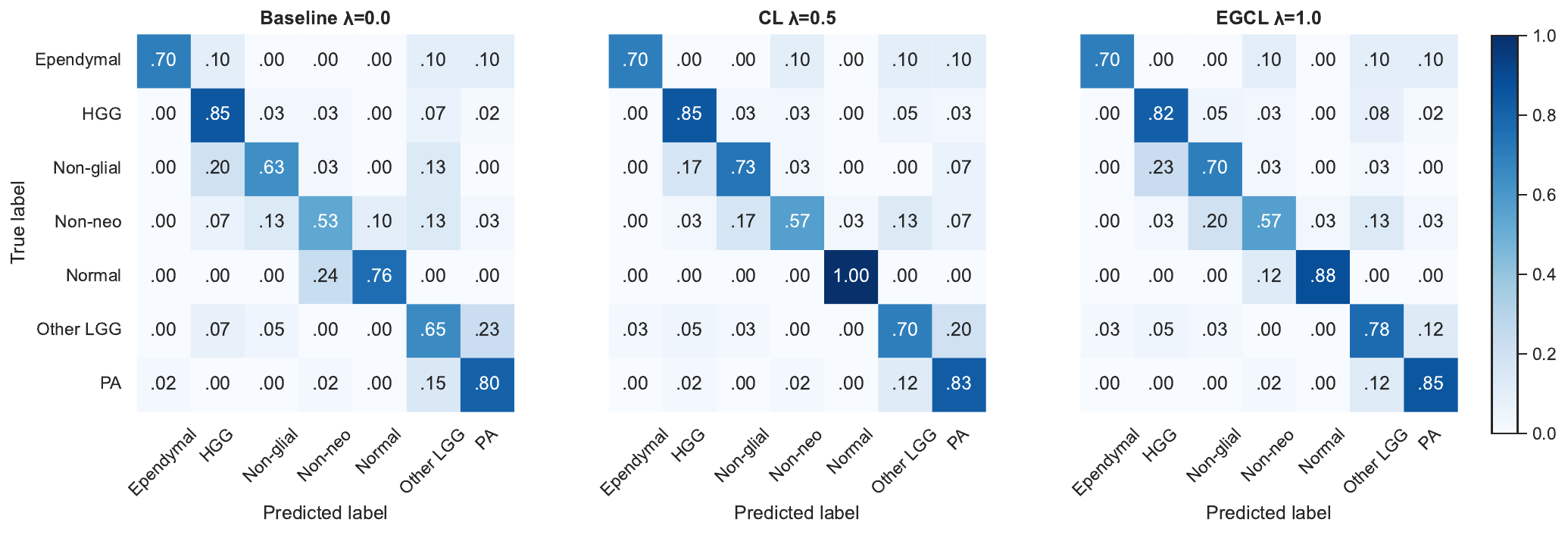}
    \caption{Normalized confusion matrices for the 7-class task comparing the baseline CLAM model ($\lambda=0$), standard supervised contrastive learning, and expert-guided contrastive learning. Contrastive objectives reduce systematic confusions between diagnostically adjacent classes while preserving strong performance on visually distinctive tumor types.}
    \label{fig:confusion_matrix}
\end{figure*}

\subsection{Classification Performance}
Table I reports 10-fold cross-validation performance for all methods across the 7-, 6-, 3-, and 2-class tasks, including per-class precision, recall, and F1-score, as well as macro-averaged metrics. Overall, the benefit of contrastive objectives is most evident in fine-grained classification: in the 7-class task, CL + CLAM achieves the best macro recall of 0.768, improving over Baseline + CLAM (0.705), with the highest per-class recall observed for Normal brain (1.000) and strong recall for High-grade brain
tumors (0.850). In the 6-class task, gains are smaller, and the best macro recall is obtained by EGCL + Linear (0.765). This setting reaches high recall on Ependymal tumors (0.900) and High-grade brain tumors (0.850), while more heterogeneous categories remain harder. In the 3-class task, CL improves macro recall to 0.658 with CL + CLAM (vs.\ 0.625 for Baseline + CLAM), driven largely by Pilocytic astrocytoma recall of 0.975. 

In contrast, contrastive learning does not yield consistent gains on the binary tumor vs.\ non-tumor task because the decision boundary is coarse and the task is already near-ceiling with only two broad categories (best macro recall 0.914 with Macenko + CLAM). The contrastive objective therefore offers limited additional supervision. The benefit of contrastive fine-tuning becomes apparent in the multi-class settings, where separating multiple confusable subtypes requires stronger embedding geometry regularization and leads to higher macro recall.
\noindent\emph{Effect of Macenko Stain Normalization}. We compared models trained on native patches to an otherwise identical pipeline using Macenko-normalized patches. In the binary tumor vs. non-tumor task (Task~1), Macenko normalization achieved the strongest performance among the evaluated preprocessing variants, suggesting that reducing stain-related variability can be beneficial for coarse tumor detection

\newcolumntype{P}[1]{>{\raggedright\arraybackslash}p{#1}}
\newcolumntype{C}[1]{>{\centering\arraybackslash}p{#1}} 

\newcommand{\slideimg}[1]{\raisebox{-\height}{\includegraphics[width=\linewidth]{#1}}}

\newcommand{\slidecell}[2]{%
\begin{minipage}[c]{\linewidth}
  \centering
  \slideimg{#1}\par
  \vspace{2pt}
  \footnotesize\texttt{#2}
\end{minipage}%
}

\begin{table*}[h]
  \centering
  \footnotesize
  \renewcommand{\arraystretch}{1.2}
  \setlength{\extrarowheight}{2.5pt}
  
\caption{Representative case-level error analysis for the 7-class task. For each slide, we show the ground-truth label and predictions from Baseline, CL, and EGCL, grouped by prediction pattern, together with expert neuropathology interpretations of likely histologic confounders.}

\BlockTitle{\textit{Baseline incorrect $\rightarrow$ CL \& EGCL correct}}
\resizebox{\textwidth}{!}{
\begin{tabular}{@{} C{1.2cm} P{2.0cm} P{2.0cm} P{2.0cm} P{2.0cm} P{14.5cm} @{}}
\toprule

\textbf{Slide (ID)} & \textbf{GT} & \textbf{Baseline} & \textbf{+CL} & \textbf{+EGCL} & \textbf{Expert's Explanation} \\
\midrule
\slidecell{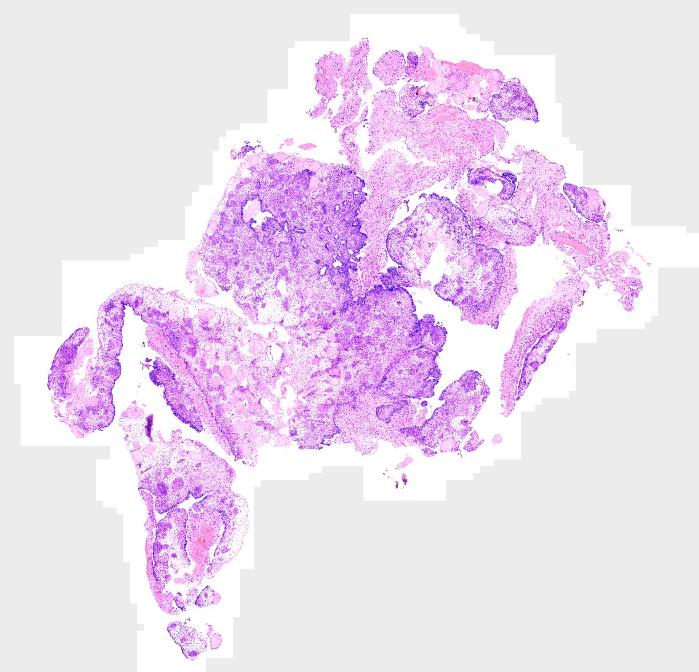}{Case 1} &
Non-glial brain tumors & Other low-grade glial tumors & Non-glial brain tumors & Non-glial brain tumors &
This histology should be very distinct from the low/high grade glial and neuronal tumors. The baseline likely latched onto superficial low-grade–like textures, while CL/EGCL better captured class-specific non-glial patterns and separated them from glial mimics.\\

\slidecell{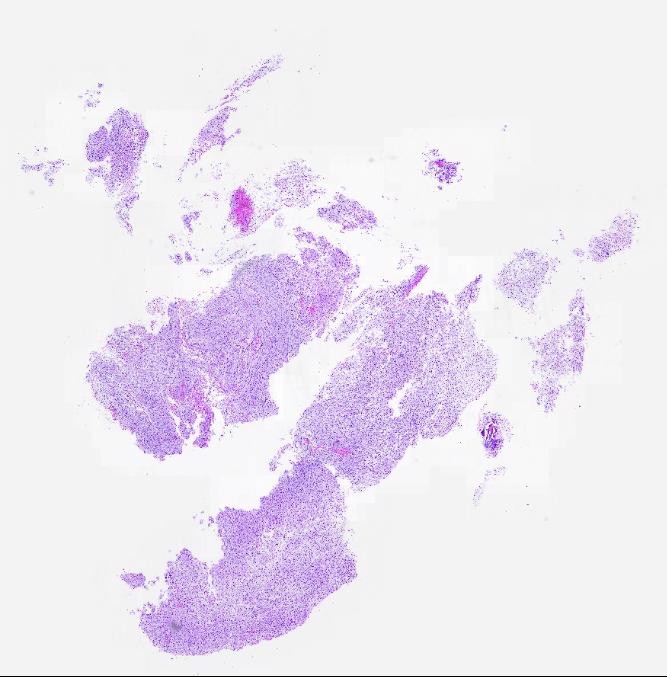}{Case 2} &
High-grade brain tumors(H3 K27M) & Pilocytic astrocytoma & High-grade brain tumors & High-grade brain tumors &
H3 K27M tumors more often overlap with other high-grade tumors, but can rarely simulate low-grade glial tumors. The baseline may have been biased by low-grade–appearing regions, whereas CL/EGCL emphasized the more global high-grade features.\\

\slidecell{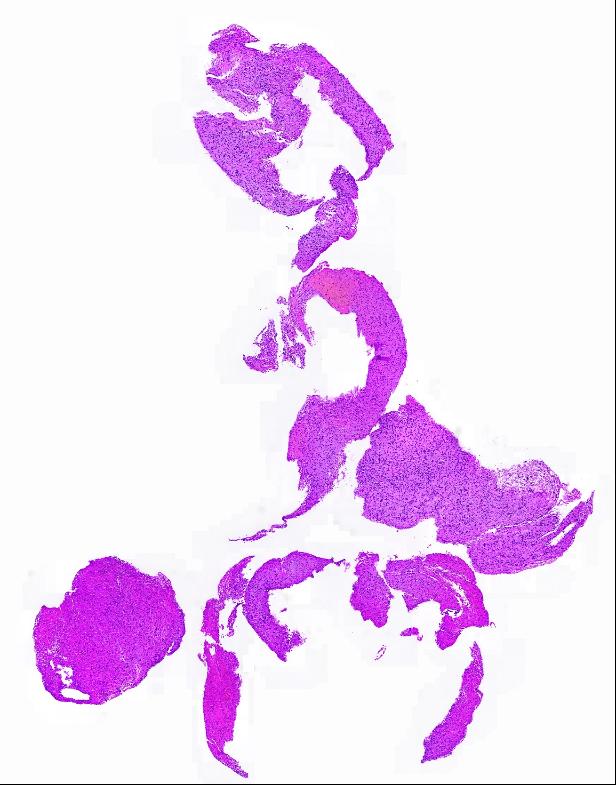}{Case 3} &
Other low-grade glial tumors & Non-glial brain tumors & Other low-grade glial tumors & Other low-grade glial tumors &
This should not be confused as a non-glial tumor, but overlap with pilocytic astro and other low-grade tumors exists. The baseline likely overreacted to non-specific morphology, while CL/EGCL reinforced finer low-grade glial cues and reduced the non-glial confusion.\\
\bottomrule
\end{tabular}
}

\vspace{4pt}

\BlockTitle{\textit{Baseline correct $\rightarrow$ CL \& EGCL incorrect}}
\resizebox{\textwidth}{!}{
\begin{tabular}{@{} C{1.2cm} P{2.0cm} P{2.0cm} P{2.0cm} P{2.0cm} P{14.5cm} @{}}
\toprule
\textbf{Slide (ID)} & \textbf{GT} & \textbf{Baseline} & \textbf{+CL} & \textbf{+EGCL} & \textbf{Expert's Explanation} \\
\midrule
\slidecell{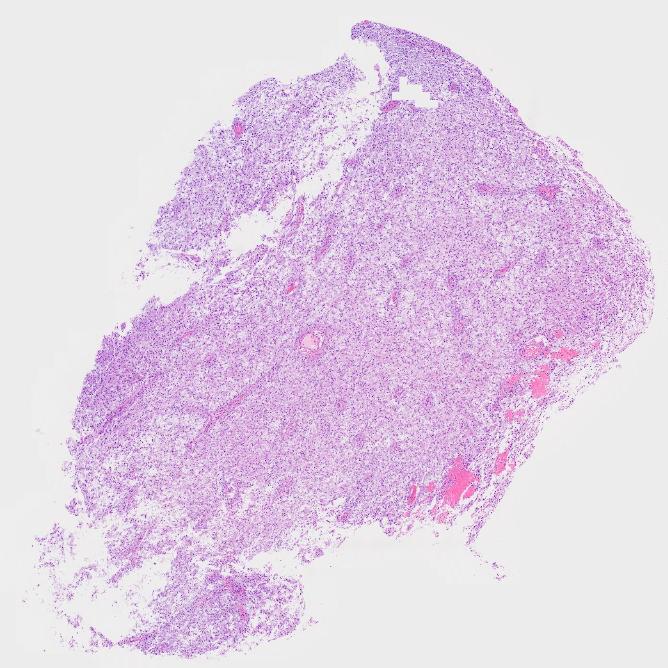}{Case 4} &
Pilocytic astrocytoma(DNET) & Pilocytic astrocytoma & High-grade brain tumors & High-grade brain tumors &
This pilocytic astrocytoma has higher cell density than typical and could overlap with the density in high-grade tumors. CL/EGCL may overweight “cellularity/density” cues, pushing this atypical PA toward the high-grade cluster.\\

\slidecell{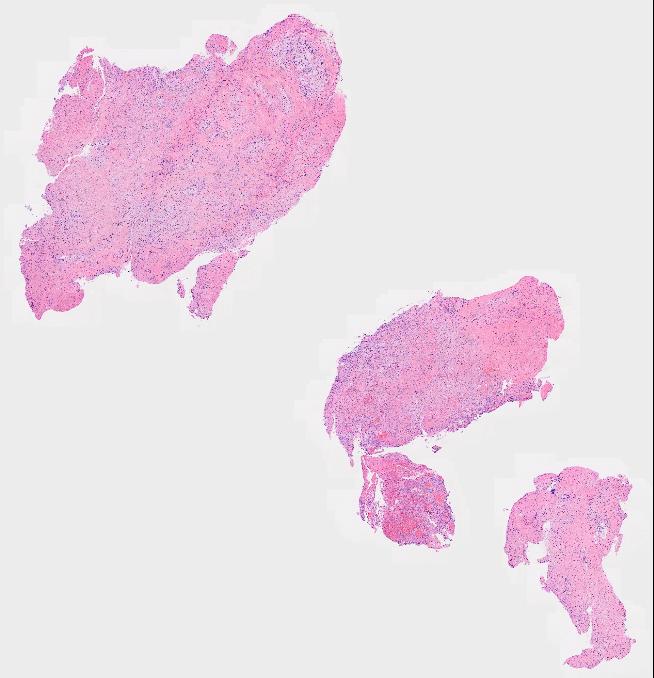}{Case 5} &
Other low-grade glial tumors & Other low-grade glial tumors & Pilocytic astrocytoma & Pilocytic astrocytoma &
This low-grade tumor (DNET) shares many overlapping features on histology with PA; even for a human it can be challenging to distinguish them. The tumor cells though have different appearance than we see with most PA.\\
\bottomrule
\end{tabular}
}

\vspace{4pt}

\BlockTitle{\textit{All incorrect}}
\resizebox{\textwidth}{!}{
\begin{tabular}{@{} C{1.2cm} P{2.0cm} P{2.0cm} P{2.0cm} P{2.0cm} P{14.5cm} @{}}
\toprule
\textbf{Slide (ID)} & \textbf{GT} & \textbf{Baseline} & \textbf{+CL} & \textbf{+EGCL} & \textbf{Expert's Explanation} \\
\midrule
\slidecell{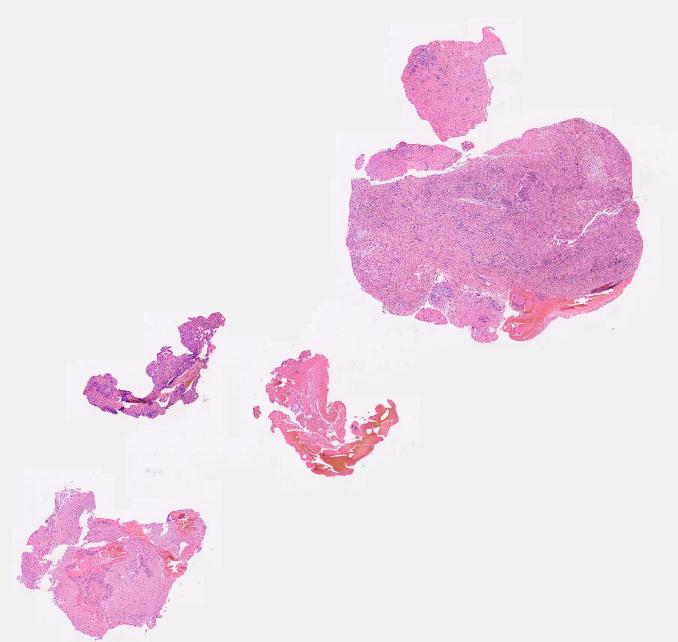}{Case 6} &
Non-neoplastic brain lesions & Other low-grade glial tumors & Other low-grade glial tumors & Other low-grade glial tumors &
Inflammation can be cellular and cause a type of reaction in normal glial cells within the brain that can simulate a low-grade brain tumor.\\

\slidecell{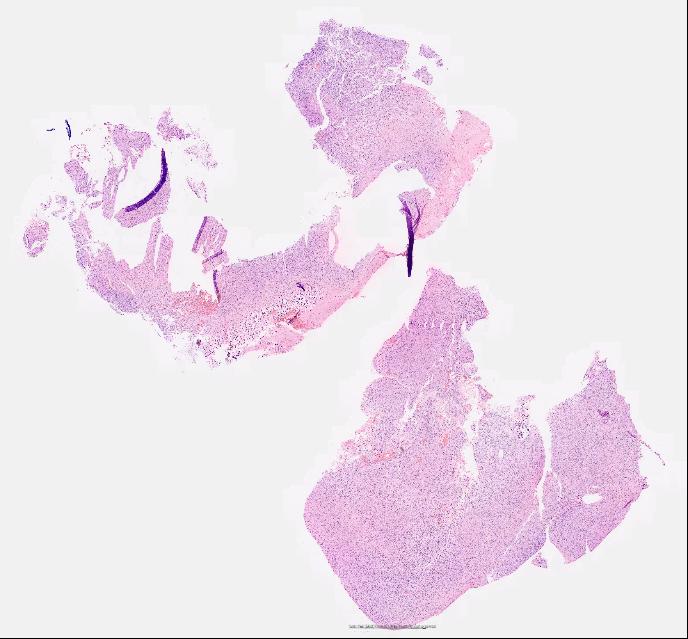}{Case 7} &
Other low-grade glial tumors & High-grade brain tumors & High-grade brain tumors & High-grade brain tumors &
This is a very tricky case. The histology overlaps very much with a high-grade tumor based on cell density and nuclear darkness, but there is not necrosis or excess mitotic figures that would be associated with high-grade. A human might consider this a high-grade tumor before getting additional information or genetics.\\

\slidecell{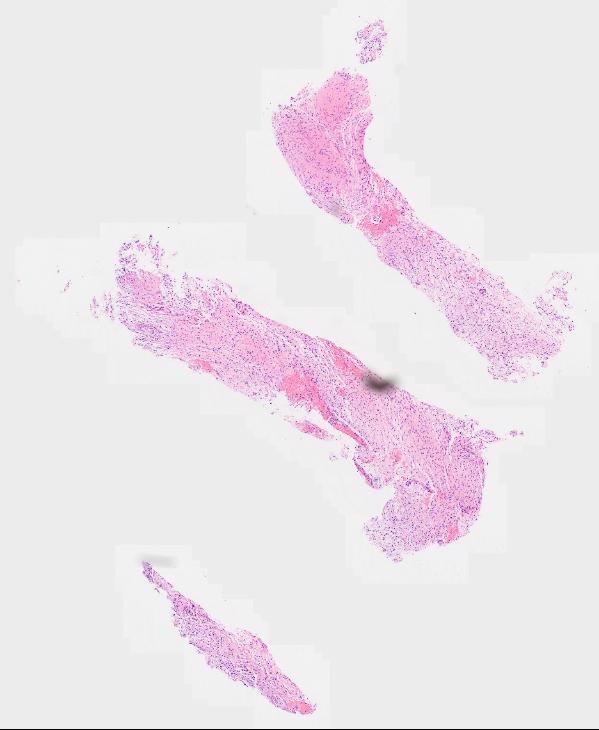}{Case 8} &
Pilocytic astrocytoma & Other low-grade glial tumors & Other low-grade glial tumors & Other low-grade glial tumors &
This is a rare subtype of pilocytic astrocytoma (pilomyxoid variant), which has significant overlap with many other low-grade glial tumors. The overlapping predictions are reasonable as a low-grade glioma is in the differential diagnosis on histologic grounds.\\
\bottomrule
\end{tabular}
}

\vspace{4pt}

\BlockTitle{\textit{Baseline incorrect, CL incorrect $\rightarrow$ EGCL correct}}
\resizebox{\textwidth}{!}{
\begin{tabular}{@{} C{1.2cm} P{2.0cm} P{2.0cm} P{2.0cm} P{2.0cm} P{14.5cm} @{}}
\toprule
\textbf{Slide (ID)} & \textbf{GT} & \textbf{Baseline} & \textbf{+CL} & \textbf{+EGCL} & \textbf{Expert's Explanation} \\
\midrule
\slidecell{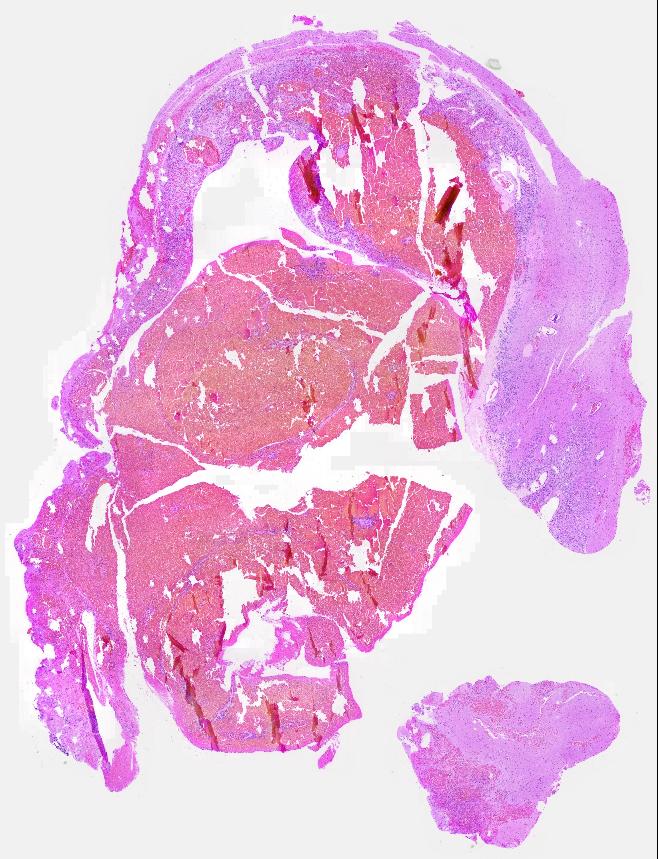}{Case 9} &
Non-glial brain tumors & Other low-grade glial tumors & Pilocytic astrocytoma & Non-glial brain tumors &
This tumor is actually one of blood vessels, but tends to recruit many reactive normal glial cells that could mimic both a pilocytic astrocytoma or different non-glial brain tumors.\\
\bottomrule
\end{tabular}
}

\vspace{4pt}

\BlockTitle{\textit{Baseline incorrect $\rightarrow$ CL correct, EGCL correct}}
\resizebox{\textwidth}{!}{
\begin{tabular}{@{} C{1.2cm} P{2.0cm} P{2.0cm} P{2.0cm} P{2.0cm} P{14.5cm} @{}}
\toprule
\textbf{Slide (ID)} & \textbf{GT} & \textbf{Baseline} & \textbf{+CL} & \textbf{+EGCL} & \textbf{Expert's Explanation} \\
\midrule
\slidecell{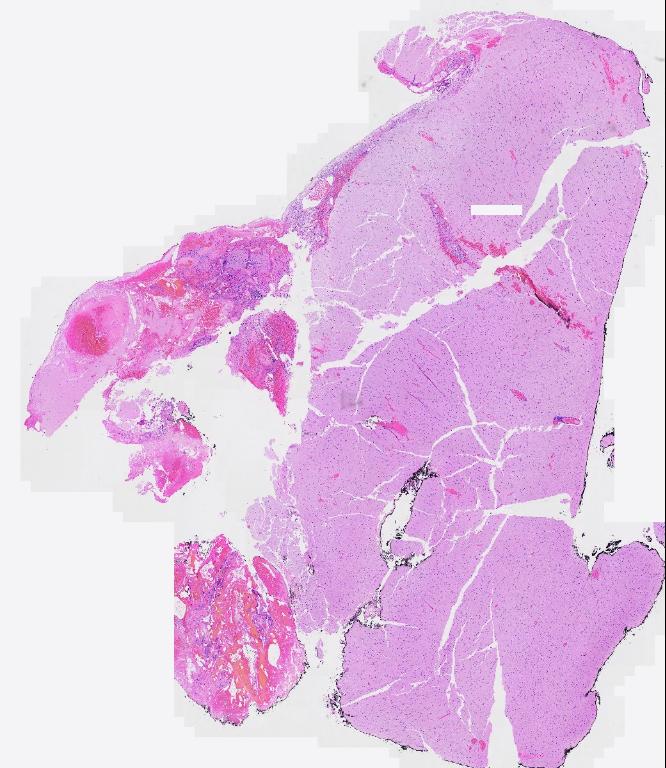}{Case 10} &
Normal brain & Non-neoplastic brain lesions & Normal brain & Non-neoplastic brain lesions &
There is always overlap between normal brain and non-neoplastic brain, especially those from epilepsy. Grouping non-neoplastic brain and normal brain is reasonable.\\
\bottomrule
\end{tabular}
}

\label{tab:error_analysis}
\end{table*}

\begin{figure}[h!]
    \centering
    \includegraphics[width=\linewidth]{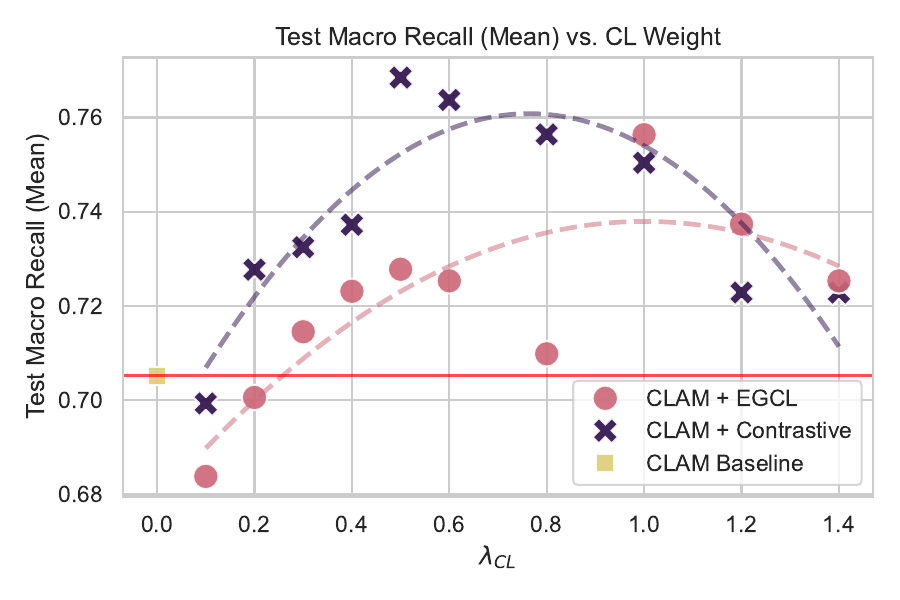}
    \caption{Sensitivity of 7-class test macro recall to the auxiliary contrastive loss weight ($\lambda$), comparing standard contrastive learning against the expert-guided variant. Performance follows an inverted-U trend, with best results at intermediate weights and degradation when the auxiliary objective dominates.}
    \label{fig:lambda_sensitivity}
\end{figure}

\subsection{Impact of Auxiliary CL Weight on Model Performance}

We evaluated the contribution of auxiliary representation learning objectives by performing a sensitivity analysis on the loss weight hyperparameter ($\lambda$) across varying classification tasks. The analysis revealed that the proposed contrastive learning approach consistently outperformed EGCL, particularly in complex multi-class settings.
In the 7-class classification task, Contrastive + CLAM followed a distinct inverted-U performance pattern shown in \textbf{Fig.~\ref{fig:lambda_sensitivity}}. Test macro recall improved as the weight increased from the baseline, peaking at $0.768$ within the range of $w=[0.3,0.6]$ (optimal $w\in0.5$). This represents a substantial improvement over the baseline performance of approximately $0.725$. At these intermediate settings, the contrastive loss provided necessary regularization to improve class separability without overwhelming the primary Multiple Instance Learning (MIL) objective. However, performance degraded at higher weights ($w>0.8$), suggesting that excessive regularization distorts the decision boundaries required for accurate classification. Conversely, EGCL proved less effective. The standard EGCL approach showed moderate improvements but remained consistently below the contrastive benchmark.
\subsection{Error Analysis and Failure Modes}
To isolate the limitations of standard MIL, we first analyzed the baseline model behavior ($w=0.0$) using per-class recall, confusion matrices, and prediction confidence distributions (\textbf{Fig.~\ref{fig:confusion_matrix}}). Performance varied substantially across diagnostic categories, reflecting differences in visual homogeneity. Classes with canonical and visually consistent patterns, such as High-grade glioma (Recall: 0.85), Pilocytic astrocytoma (0.80), and Normal brain (0.76), were classified with relatively high accuracy. In contrast, ``umbrella'' or heterogeneous categories, including Non-neoplastic lesions (0.53) and Non-glial tumors (0.63), exhibited substantially lower recall, indicating intrinsic difficulty in defining a unified feature representation for these classes.

Confusion matrices revealed that errors were biologically meaningful rather than random. For Normal Brain, 24\% of slides (4/17) were misclassified as Non-neoplastic lesions, highlighting the difficulty in distinguishing reactive tissue from true normal parenchyma based on morphology alone. Similarly, Non-neoplastic lesions were frequently dispersed across neoplastic categories (misclassified as Non-glial or Other LGG), consistent with the model over-relying on cellular density rather than cell identity.

\section{Discussion}
\textbf{Case-Level Error Analysis with Expert Interpretation}
To contextualize these findings, we conducted a targeted, expert-guided analysis comparing the baseline predictions to Contrastive Learning (CL) variants (Table~\ref{tab:error_analysis}).
Correction of Superficial Mimics: In several cases where the baseline failed but CL variants succeeded, errors were attributable to superficial morphologic overlap. This is a known challenge for experienced pathologists. Many critical histopathologic features (tumor cell density, nuclear hyperchromasia, presence of tumor necrosis or vascular proliferation) can partially exist across disparate classes. For example, Case 1 (class non-glial tumor,  represented by a craniopharyngioma) was misclassified by the baseline as a low-grade glioma.  Possible explanations for this mis-classification could be the similar cellular density and size of tumor nuclei among this particular non-glial tumor and many types of low-grade gliomas. In general, it is understood that lower-grade brain tumors and higher grade brain tumors exhibit a continuum of these critical histopathologic features without clear dividing boundaries.  In contrast to baseline, the CL models correctly identified the non-glial patterns, suggesting the auxiliary loss enforces sensitivity to class-specific global features rather than isolated local cues.

\textbf{New Modes of Confusion:} Conversely, contrastive learning introduced specific trade-offs. In cases such as Case 4 (class other low-grade glioma, represented as a dysembryoplastic neuroepithelial tumor) , the tumor displayed an unusually high cellularity compared to what is typically observed for this particular type of low-grade glioma. While the baseline correctly identified it, the CL-based models misclassified it as high-grade glioma. This suggests that by explicitly pushing representations apart, the contrastive objective may overweight features like ``cellular density,'' resulting in assignment of a densely cellular low-grade tumor toward the high-grade cluster. The assignment may also discount other important histopathologic features that could mitigate such an assignment.

\textbf{Intrinsic Ambiguity:} A subset of cases remained incorrect across all models, illustrating the ``Bayes error'' limit of H\&E histology. Case 6 (class non-neoplastic lesion, represented as non-neoplastic inflamed brain tissue)  illustrates a tricky manner in which the inflammatory process within the brain tissue exhibits sufficient cell density to superficially  mimic a low-grade tumor. It is possible that, once again, over-weighting of the cellular density feature resulted in a similar type of misclassification error across all models. Similarly, Case 7 (class Other LGG, represented as a low-grade glioneuronal tumor ) displayed both increased cell density and atypical nuclear features among the tumor cells. Alone, these two histopathologic characteristics would lead expert neuropathologists to reasonably consider classification among high-grade gliomas. Some mitigating features arguing against this classification would be the  lack of tumor necrosis. Nevertheless, the baseline histopathology shares significant overlap with the histopathology in many high-grade gliomas. Adjudication of such cases with diagnostic class overlap often requires integration of both clinical context (age of patient, tumor location in the brain) or integration of ancillary information from genetic testing or immunohistochemistry profiles. These specific errors reinforce the need to expand the histopathologic feature-set beyond those recognized by human experience, in order to transcend the need to rely on molecular-genetic information or other ancillary technique. This is true regardless of the representation learning strategy.

\section*{Acknowledgment}
This work was supported by the NSF AI Institute for Foundations of Machine Learning. The experiments were conducted on the Vista GPU cluster through the Center for Generative AI (CGAI) and the Texas Advanced Computing Center (TACC) at The University of Texas at Austin. This research was also partially funded by the National Institutes of Health (NIH) under award 1R01EB03710101. The views and conclusions contained in this document are those of the authors and should not be interpreted as representing the official policies, either expressed or implied, of the NIH.
\bibliographystyle{ieeetr}
\bibliography{main}

@misc{xiongSurveyPathologyFoundation2025,
  title = {A {{Survey}} of {{Pathology Foundation Model}}: {{Progress}} and {{Future Directions}}},
  shorttitle = {A {{Survey}} of {{Pathology Foundation Model}}},
  author = {Xiong, Conghao and Chen, Hao and Sung, Joseph J. Y.},
  year = 2025,
  month = apr,
  publisher = {arXiv},
  doi = {10.48550/ARXIV.2504.04045},
  urldate = {2025-05-25},
  copyright = {Creative Commons Attribution 4.0 International}
}

@article{xiangVisionLanguageFoundation2025,
  title = {A Vision--Language Foundation Model for Precision Oncology},
  author = {Xiang, Jinxi and Wang, Xiyue and Zhang, Xiaoming and Xi, Yinghua and Eweje, Feyisope and Chen, Yijiang and Li, Yuchen and Bergstrom, Colin and Gopaulchan, Matthew and Kim, Ted and Yu, Kun-Hsing and Willens, Sierra and Olguin, Francesca Maria and Nirschl, Jeffrey J. and Neal, Joel and Diehn, Maximilian and Yang, Sen and Li, Ruijiang},
  year = 2025,
  month = feb,
  journal = {Nature},
  volume = {638},
  number = {8051},
  pages = {769--778},
  issn = {0028-0836, 1476-4687},
  doi = {10.1038/s41586-024-08378-w},
  urldate = {2025-10-09},
  langid = {english}
}

@misc{vadoriMindGapEvaluating2025,
  title = {Mind the {{Gap}}: {{Evaluating Patch Embeddings}} from {{General-Purpose}} and {{Histopathology Foundation Models}} for {{Cell Segmentation}} and {{Classification}}},
  shorttitle = {Mind the {{Gap}}},
  author = {Vadori, Valentina and Peruffo, Antonella and Gra{\"i}c, Jean-Marie and Finos, Livio and Grisan, Enrico},
  year = 2025,
  month = feb,
  publisher = {arXiv},
  doi = {10.48550/ARXIV.2502.02471},
  urldate = {2025-06-24},
  copyright = {Creative Commons Attribution Non Commercial Share Alike 4.0 International}
}

@article{tampuPediatricBrainTumor2025,
  title = {Pediatric Brain Tumor Classification Using Digital Pathology and Deep Learning: {{Evaluation}} of {{{\textsc{SOTA}}}} Methods on a Multi-center {{Swedish}} Cohort},
  shorttitle = {Pediatric Brain Tumor Classification Using Digital Pathology and Deep Learning},
  author = {Tampu, Iulian Emil and Nyman, Per and Spyretos, Christoforos and Blystad, Ida and Shamikh, Alia and Prochazka, Gabriela and De St{\aa}hl, Teresita D{\'i}az and Sandgren, Johanna and Lundberg, Peter and Haj-Hosseini, Neda},
  year = 2025,
  month = jun,
  journal = {Brain Pathology},
  pages = {e70029},
  issn = {1015-6305, 1750-3639},
  doi = {10.1111/bpa.70029},
  urldate = {2025-11-13},
  langid = {english}
}

@article{saueressigHistologyDiagnosisLeveraging2025,
  title = {From Histology to Diagnosis: {{Leveraging}} Pathology Foundation Models for Glioma Classification},
  shorttitle = {From Histology to Diagnosis},
  author = {Saueressig, Camillo and Delbridge, Claire and Scholz, Daniel and Kazemi, Azar and Khan, Mohammad Zaid and Metz, Marie and Meyer, Bernhard and Mitsdoerffer, Meike and Sch{\"u}ffler, Peter J. and Wiestler, Benedikt},
  year = 2025,
  month = aug,
  journal = {Computers in Biology and Medicine},
  volume = {197},
  pages = {110988},
  issn = {00104825},
  doi = {10.1016/j.compbiomed.2025.110988},
  urldate = {2025-12-11},
  langid = {english}
}

@misc{liCanWeSimplify2025,
  title = {Can {{We Simplify Slide-level Fine-tuning}} of {{Pathology Foundation Models}}?},
  author = {Li, Jiawen and Hu, Jiali and Sun, Qiehe and Yan, Renao and Ouyang, Minxi and Guan, Tian and Han, Anjia and He, Chao and He, Yonghong},
  year = 2025,
  month = feb,
  publisher = {arXiv},
  doi = {10.48550/ARXIV.2502.20823},
  urldate = {2025-11-10},
  copyright = {arXiv.org perpetual, non-exclusive license}
}

@misc{karasikovTrainingStateoftheartPathology2025,
  title = {Training State-of-the-Art Pathology Foundation Models with Orders of Magnitude Less Data},
  author = {Karasikov, Mikhail and {van Doorn}, Joost and K{\"a}nzig, Nicolas and Cesur, Melis Erdal and Horlings, Hugo Mark and Berke, Robert and Tang, Fei and Ot{\'a}lora, Sebastian},
  year = 2025,
  month = apr,
  publisher = {arXiv},
  doi = {10.48550/ARXIV.2504.05186},
  urldate = {2025-10-09},
  copyright = {arXiv.org perpetual, non-exclusive license}
}

@misc{gSequentialAttentionbasedSampling2025,
  title = {Sequential {{Attention-based Sampling}} for {{Histopathological Analysis}}},
  author = {G, Tarun and Malpani, Naman and Thoppe, Gugan and Devarajan, Sridharan},
  year = 2025,
  month = jul,
  publisher = {arXiv},
  doi = {10.48550/ARXIV.2507.05077},
  urldate = {2025-12-11},
  copyright = {Creative Commons Attribution Non Commercial No Derivatives 4.0 International}
}

@article{dingMultimodalWholeslideFoundation2025,
  title = {A Multimodal Whole-Slide Foundation Model for Pathology},
  author = {Ding, Tong and Wagner, Sophia J. and Song, Andrew H. and Chen, Richard J. and Lu, Ming Y. and Zhang, Andrew and Vaidya, Anurag J. and Jaume, Guillaume and Shaban, Muhammad and Kim, Ahrong and Williamson, Drew F. K. and Robertson, Harry and Chen, Bowen and {Almagro-P{\'e}rez}, Cristina and Doucet, Paul and Sahai, Sharifa and Chen, Chengkuan and Chen, Christina S. and Komura, Daisuke and Kawabe, Akihiro and Ochi, Mieko and Sato, Shinya and Yokose, Tomoyuki and Miyagi, Yohei and Ishikawa, Shumpei and Gerber, Georg and Peng, Tingying and Le, Long Phi and Mahmood, Faisal},
  year = 2025,
  month = nov,
  journal = {Nature Medicine},
  issn = {1078-8956, 1546-170X},
  doi = {10.1038/s41591-025-03982-3},
  urldate = {2025-11-13},
  langid = {english}
}

@article{zhangPatchesWSIsSystematic2024,
  title = {From Patches to {{WSIs}}: {{A}} Systematic Review of Deep {{Multiple Instance Learning}} in Computational Pathology},
  shorttitle = {From Patches to {{WSIs}}},
  author = {Zhang, Yuchen and Gao, Zeyu and He, Kai and Li, Chen and Mao, Rui},
  year = 2024,
  month = dec,
  journal = {Information Fusion},
  volume = {119},
  pages = {103027},
  issn = {15662535},
  doi = {10.1016/j.inffus.2025.103027},
  urldate = {2025-07-29},
  langid = {english}
}

@misc{tampuPediatricBrainTumor2024a,
  title = {Pediatric Brain Tumor Classification Using Digital Histopathology and Deep Learning: Evaluation of {{SOTA}} Methods on a Multi-Center {{Swedish}} Cohort},
  shorttitle = {Pediatric Brain Tumor Classification Using Digital Histopathology and Deep Learning},
  author = {Tampu, Iulian Emil and Nyman, Per and Spyretos, Christoforos and Blystad, Ida and Shamikh, Alia and Prochazka, Gabriela and {de St{\aa}hl}, Teresita D{\'i}az and Sandgren, Johanna and Lundberg, Peter and {Haj-Hosseini}, Neda},
  year = 2024,
  month = sep,
  publisher = {arXiv},
  doi = {10.48550/ARXIV.2409.01330},
  urldate = {2025-05-25},
  copyright = {Creative Commons Attribution 4.0 International}
}

@article{shiPositionalEncodingguidedTransformerbased2024,
  title = {Positional Encoding-Guided Transformer-Based Multiple Instance Learning for Histopathology Whole Slide Images Classification},
  author = {Shi, Jun and Sun, Dongdong and Wu, Kun and Jiang, Zhiguo and Kong, Xue and Wang, Wei and Wu, Haibo and Zheng, Yushan},
  year = 2024,
  month = jun,
  journal = {Computer Methods and Programs in Biomedicine},
  volume = {258},
  pages = {108491},
  publisher = {Elsevier BV},
  issn = {0169-2607},
  doi = {10.1016/j.cmpb.2024.108491},
  urldate = {2025-07-10},
  copyright = {https://www.elsevier.com/tdm/userlicense/1.0/},
  langid = {english}
}

@misc{neidlingerBenchmarkingFoundationModels2024,
  title = {Benchmarking Foundation Models as Feature Extractors for Weakly-Supervised Computational Pathology},
  author = {Neidlinger, Peter and Nahhas, Omar S. M. El and Muti, Hannah Sophie and Lenz, Tim and Hoffmeister, Michael and Brenner, Hermann and {van Treeck}, Marko and Langer, Rupert and Dislich, Bastian and Behrens, Hans Michael and R{\"o}cken, Christoph and Foersch, Sebastian and Truhn, Daniel and Marra, Antonio and Saldanha, Oliver Lester and Kather, Jakob Nikolas},
  year = 2024,
  month = dec,
  publisher = {arXiv},
  doi = {10.48550/ARXIV.2408.15823},
  urldate = {2025-07-29},
  copyright = {Creative Commons Attribution 4.0 International}
}

@article{chenGeneralpurposeFoundationModel2023,
  title = {Towards a General-Purpose Foundation Model for Computational Pathology},
  author = {Chen, Richard J. and Ding, Tong and Lu, Ming Y. and Williamson, Drew F. K. and Jaume, Guillaume and Song, Andrew H. and Chen, Bowen and Zhang, Andrew and Shao, Daniel and Shaban, Muhammad and Williams, Mane and Oldenburg, Lukas and Weishaupt, Luca L. and Wang, Judy J. and Vaidya, Anurag and Le, Long Phi and Gerber, Georg and Sahai, Sharifa and Williams, Walt and Mahmood, Faisal},
  year = 2023,
  month = aug,
  journal = {Nature Medicine},
  volume = {30},
  number = {3},
  pages = {850--862},
  issn = {1078-8956, 1546-170X},
  doi = {10.1038/s41591-024-02857-3},
  urldate = {2025-05-25},
  langid = {english}
}

@article{tampu2025pediatric,
  title={Pediatric brain tumor classification using deep learning on MR images with age fusion},
  author={Tampu, Iulian Emil and Bianchessi, Tamara and Blystad, Ida and Lundberg, Peter and Nyman, Per and Eklund, Anders and Haj-Hosseini, Neda},
  journal={Neuro-Oncology Advances},
  volume={7},
  number={1},
  pages={vdae205},
  year={2025},
  publisher={Oxford University Press US}
}

@misc{he2020momentumcontrastunsupervisedvisual,
      title={Momentum Contrast for Unsupervised Visual Representation Learning}, 
      author={Kaiming He and Haoqi Fan and Yuxin Wu and Saining Xie and Ross Girshick},
      year={2020},
      eprint={1911.05722},
      archivePrefix={arXiv},
      primaryClass={cs.CV},
      url={https://arxiv.org/abs/1911.05722}, 
}

@article{dorfner2025review,
  title={A review of deep learning for brain tumor analysis in MRI},
  author={Dorfner, Felix J and Patel, Jay B and Kalpathy-Cramer, Jayashree and Gerstner, Elizabeth R and Bridge, Christopher P},
  journal={NPJ Precision Oncology},
  volume={9},
  number={1},
  pages={2},
  year={2025},
  publisher={Nature Publishing Group UK London}
}

@inproceedings{ilse2018attention,
  title={Attention-based deep multiple instance learning},
  author={Ilse, Maximilian and Tomczak, Jakub and Welling, Max},
  booktitle={International conference on machine learning},
  pages={2127--2136},
  year={2018},
  organization={PMLR}
}

@inproceedings{chen2020simple,
  title={A simple framework for contrastive learning of visual representations},
  author={Chen, Ting and Kornblith, Simon and Norouzi, Mohammad and Hinton, Geoffrey},
  booktitle={International conference on machine learning},
  pages={1597--1607},
  year={2020},
  organization={PmLR}
}

@article{tampu2026pediatric,
  title={Pediatric brain tumor classification using digital pathology and deep learning: Evaluation of SOTA methods on a multi-center Swedish cohort},
  author={Tampu, Iulian Emil and Nyman, Per and Spyretos, Christoforos and Blystad, Ida and Shamikh, Alia and Prochazka, Gabriela and de St{\aa}hl, Teresita D{\'\i}az and Sandgren, Johanna and Lundberg, Peter and Haj-Hosseini, Neda},
  journal={Brain Pathology},
  volume={36},
  number={1},
  pages={e70029},
  year={2026},
  publisher={Wiley Online Library}
}

@article{viaene2023pediatric,
  title={Pediatric brain tumors: A neuropathologist's approach to the integrated diagnosis},
  author={Viaene, Angela N},
  journal={Frontiers in Pediatrics},
  volume={11},
  pages={1143363},
  year={2023},
  publisher={Frontiers Media SA}
}

@article{price2025cbtrus,
  title={CBTRUS statistical report: Pediatric brain tumor foundation childhood and adolescent primary brain and other central nervous system tumors diagnosed in the United States in 2017-2021},
  author={Price, Mackenzie and Ballard, Christine Ann Pittman and Benedetti, Julia R and Kruchko, Carol and Barnholtz-Sloan, Jill S and Ostrom, Quinn T},
  journal={Neuro-Oncology},
  volume={27},
  number={Supplement\_1},
  pages={i1--i42},
  year={2025},
  publisher={Oxford University Press US}
}

@article{louis20212021,
  title={The 2021 WHO classification of tumors of the central nervous system: a summary},
  author={Louis, David N and Perry, Arie and Wesseling, Pieter and Brat, Daniel J and Cree, Ian A and Figarella-Branger, Dominique and Hawkins, Cynthia and Ng, HK and Pfister, Stefan M and Reifenberger, Guido and others},
  journal={Neuro-oncology},
  volume={23},
  number={8},
  pages={1231--1251},
  year={2021},
  publisher={Oxford University Press US}
}

@article{zhang2025standardizing,
  title={Accelerating Data Processing and Benchmarking of AI Models for Pathology},
  author={Zhang, Andrew and Jaume, Guillaume and Vaidya, Anurag and Ding, Tong and Mahmood, Faisal},
  journal={arXiv preprint arXiv:2502.06750},
  year={2025}
}

@article{vaidya2025molecular,
  title={Molecular-driven Foundation Model for Oncologic Pathology},
  author={Vaidya, Anurag and Zhang, Andrew and Jaume, Guillaume and Song, Andrew H and Ding, Tong and Wagner, Sophia J and Lu, Ming Y and Doucet, Paul and Robertson, Harry and Almagro-Perez, Cristina and others},
  journal={arXiv preprint arXiv:2501.16652},
  year={2025}
}

@article{jaume2024hest,
    author = {Jaume, Guillaume and Doucet, Paul and Song, Andrew H. and Lu, Ming Y. and Almagro-Perez, Cristina and Wagner, Sophia J. and Vaidya, Anurag J. and Chen, Richard J. and Williamson, Drew F. K. and Kim, Ahrong and Mahmood, Faisal},
    title = {{HEST-1k: A Dataset for Spatial Transcriptomics and Histology Image Analysis}},
    journal = {arXiv},
    year = {2024},
    month = jun,
    eprint = {2406.16192},
    url = {https://arxiv.org/abs/2406.16192v1}
}

@inproceedings{Macenko2009AMF,
  title={A method for normalizing histology slides for quantitative analysis},
  author={Marc Macenko and Marc Niethammer and J. S. Marron and David Borland and John T. Woosley and Xiaojun Guan and Charles Schmitt and Nancy E. Thomas},
  booktitle={2009 IEEE International Symposium on Biomedical Imaging: From Nano to Macro},
  year={2009},
  pages={1107-1110}
}

@article{lu2021clam,
  title={Data-efficient and weakly supervised computational pathology on whole-slide images},
  author={Lu, Ming Y and Williamson, Drew FK and Chen, Tiffany Y and Chen, Richard J and Barbieri, Matteo and Mahmood, Faisal},
  journal={Nature Biomedical Engineering},
  volume={5},
  number={6},
  pages={555--570},
  year={2021},
  publisher={Nature Publishing Group},
  doi={10.1038/s41551-020-00682-w}
}

@article{oord2018representation,
  title={Representation Learning with Contrastive Predictive Coding},
  author={van den Oord, A{\"a}ron and Li, Yazhe and Vinyals, Oriol},
  journal={arXiv preprint arXiv:1807.03748},
  year={2018}
}

@inproceedings{khosla2020supervised,
  title={Supervised Contrastive Learning},
  author={Khosla, Prannay and Teterwak, Piotr and Wang, Chen and Sarna, Aaron and Tian, Yonglong and Isola, Phillip and Maschinot, Aaron and Liu, Ce and Krishnan, Dilip},
  booktitle={Advances in Neural Information Processing Systems},
  volume={33},
  pages={18661--18673},
  year={2020}
}

\end{document}